\documentclass{article} 
\usepackage{iclr2027_conference,times}

\usepackage{amsmath,amsfonts,bm}

\def\eqref#1{equation~\ref{#1}}

\def\1{\bm{1}}

\DeclareMathAlphabet{\mathsfit}{\encodingdefault}{\sfdefault}{m}{sl}
\SetMathAlphabet{\mathsfit}{bold}{\encodingdefault}{\sfdefault}{bx}{n}

\usepackage{bm}
\usepackage{hyperref}
\usepackage{url}
\usepackage{array}
\usepackage{hyperref}
\usepackage{url}
\usepackage{graphicx} 
\usepackage{wrapfig}
\usepackage{booktabs}
\usepackage{float}
\usepackage{placeins}
\usepackage{caption}
\usepackage{multirow}
\usepackage{wrapfig}
\usepackage{makecell}
\usepackage{colortbl}   
\usepackage{xcolor}     
\usepackage{amssymb}
\usepackage{multirow}
\usepackage{hyperref}
\usepackage[normalem]{ulem}

\title{Atomizer-IO: Beyond Pixels, Patches and Grids}

\author{
\textbf{\href{https://openreview.net/profile?id=\%7EHugo_Riffaud_de_Turckheim1}
{Hugo Riffaud de Turckheim}}$^{1}$,
\textbf{\href{https://openreview.net/profile?id=\%7ESylvain_Lobry1}
{Sylvain Lobry}}$^{2}$,
\textbf{\href{https://openreview.net/profile?id=\%7ENicolas_Houdr\%C3\%A91}
{Nicolas Houdré}}$^{2}$, \\
\textbf{\href{https://openreview.net/profile?id=\%7EDamien_Robert1}
{Damien Robert}}$^{3}$,
\textbf{\href{https://openreview.net/profile?id=\%7ERoberto_Interdonato1}
{Roberto Interdonato}}$^{4}$,
\textbf{\href{https://openreview.net/profile?id=\%7EDiego_Marcos1}
{Diego Marcos}}$^{1}$ \\[1ex]
\normalfont
$^{1}$INRIA, Montpellier, France \\
$^{2}$LIPADE, Université Paris Cité, Paris, France \\
$^{3}$DM3L, University of Zurich, Zurich, Switzerland \\
$^{4}$CIRAD, Montpellier, France \\
\texttt{hugo.riffaud--de-turckheim@inria.fr}
}

\iclrfinalcopy 
\begin{document}

\maketitle

\begin{abstract}
Most vision architectures assume that observations lie on a regular grid, an effective abstraction for natural images but a restrictive one for sensing data whose channels, temporal sampling, spatial resolution, and geometry can vary.
Generic set-based architectures remove the grid, but also remove useful spatial inductive biases. We introduce Atomizer-IO, an architecture that places observations first and derives structure from their physical relationships.
Building on top of an atomic representation of the data, each observation is described by
its measurement and acquisition metadata, while local cross-attention maps observations to anchor points that can be arbitrarily placed. We evaluate this design by progressively relaxing the grid assumption, from
varying input raster configurations and incomplete channel sets to flexible output density and, ultimately, inputs without a raster grid. Atomizer-IO is competitive with flexible EO-specific architectures on most tasks, while offering post-training control over inference cost and competitive compute--performance trade-offs. The same formulation extends without architectural redesign to
unordered 3D point clouds, showing that the atomic interface generalizes beyond regular raster inputs. These results suggest that pixels, patches, and grids do not need to define the interface of a sensing architecture.
\end{abstract}

\vspace{-0.3cm}
\section{Introduction}
\vspace{-0.3cm}

Most computer vision models, including convolutional neural networks (CNN) \citep{lecun1989cnn} and vision transformers (ViT) \citep{dosovitskiy2020image}, see the world as a regular grid. They operate on tensors whose channels and spatial organization are known in advance. Natural images fit this interface well: RGB channels have fixed semantics, pixels lie on a regular lattice, and neighboring entries have a constant spatial offset. Convolutions exploit this structure explicitly, while ViTs retain the same regular input organization even as they relax locality. In many sensing domains, however, spectral content, spatial sampling, acquisition time, and even geometry vary, making a fixed tensor overly restrictive.

Earth observation (EO) is an acute instance of this mismatch. Data in this field is heterogeneous by nature, since different sensors observe different physical quantities, at different resolutions, and at different times. Channel position therefore provides no stable physical identity across sensors. A short-wave infrared measurement sensitive to moisture and a SAR backscatter measurement sensitive to surface roughness may occupy analogous tensor positions while
representing fundamentally different physical quantities. In standard architectures, measurement identity is therefore encoded implicitly through channel position rather than explicitly through physical characteristics \citep{jakubik2025terramind}. Measurement identity is only one axis: resolution, acquisition time, channel availability, and geometry vary too.

Recent EO architectures increasingly accommodate these sources of variation through dedicated mechanisms. Rather than changing the underlying representation, they typically adapt a tensor- or patch-based interface to each source of variation. Within this fixed interface, a missing channel must be padded \citep{sumbul2025smarties} with values the model cannot distinguish from real measurements, differing resolutions resampled or handled by resolution-aware processing \citep{marsocci2024pangaea}, and irregular observations forced onto a grid before entering the model ~\citep{weinstein2021benchmark}.
Each mechanism compensates for an assumption introduced by the representation itself.

Our proposed Atomizer-IO reverses this order: \emph{observations come first, structure second}. Each observation is represented independently by its measurement and acquisition metadata, without requiring a fixed channel count, tensor shape, or neighboring measurement. Structure is introduced afterwards from physical relationships between observations: locality and relative geometry are defined directly in physical coordinates rather than inherited from tensor position.

Atomizer-IO implements this principle through an atomic encoder--processor--decoder architecture. Following Atomizer~\citep{Turckheim_2025_BMVC}, each observation is represented as an \emph{atom}; for raster data, one atom corresponds to one measurement from one spectral
band, at one pixel location and one acquisition time. The encoder aggregates atoms into spatially anchored latent vectors, the processor lets these latents interact, and a query-based decoder maps local latent neighborhoods to requested output locations. Because atomic granularity can produce millions of observations per scene, each latent attends only to a local neighborhood in physical space, while the number of spatial latents scales with the number of input observations. The same local aggregation serves as both input and output interface, allowing channel sets, input resolution, and output density to vary while extending the same architecture from regular rasters to unordered 3D point clouds.

\begin{figure}[t!]
  \centering
  \vspace{-0.25cm}
  \includegraphics[width=1.0\textwidth]{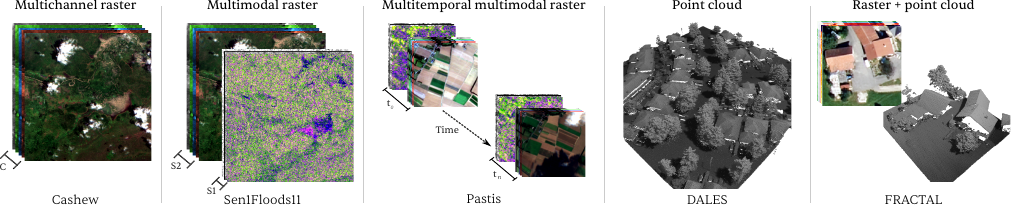}
  \vspace{-0.55cm}
\caption{
Sensing configurations considered in this work. All are processed through the
same atomic interface, without rasterization, resampling, or modality-specific
branches.
}
  \label{fig:Atomizer_sah}
  \vspace{-0.45cm}
\end{figure}

We evaluate this design by progressively relaxing the assumptions of a fixed grid. We first assess competitiveness on conventional raster EO tasks, then vary available channels and output-query density, and finally remove the raster assumption with unordered LiDAR and joint image--point observations.

Our main contributions are:
\begin{itemize}
    \item \textbf{A physically grounded atomic representation} in which
measurement identity is described explicitly rather than tied to tensor position. A single trained model spans sensor configurations that a fixed-channel architecture can only reach by padding or redesign.

    \item \textbf{A scalable, structured set-to-latent architecture} that uses the
same local geometric operation to encode observations and decode arbitrary
query locations, removing the need for a fixed grid on either side.

    \item \textbf{A single architecture across raster and irregular sensing
    geometries}, spanning classification, regression, 2D and 3D segmentation,
    and optical, radar, and LiDAR observations. All models are trained from
    scratch under matched conditions, isolating architectural differences from
    pretraining.
\end{itemize}

\vspace{-0.3cm}
\section{Related Work}
\vspace{-0.3cm}

Regular grids supply useful spatial structure, including locality, stable neighborhoods, and regular geometric relationships, that grid-based architectures can exploit directly. In Earth observation, the representation must additionally accommodate several physical axes: spatial resolution determines how space is sampled, spectral bands sample the electromagnetic spectrum, and acquisition times sample an evolving scene. 

Recent EO models increasingly expose these physical properties to grid-based
architectures through dedicated mechanisms. DOFA~\citep{xiong2024neural},
SenPa-MAE~\citep{prexl2024senpa}, and Panopticon~\citep{waldmann2025panopticon}
primarily address spectral variation, while FlexiMo~\citep{li2025fleximo}
additionally handles spatial variation but does not natively model temporal
sequences. ScaleMAE~\citep{reed2023scale} incorporates ground sampling distance,
AnySat~\citep{astruc2024anysat} defines patches in physical rather than pixel
units, and Galileo~\citep{tseng2025galileo} and
Prithvi~\citep{szwarcman2025prithvi} introduce temporal structure.
RAMEN~\citep{Houdre_2026_CVPR} and UniverSat~\citep{perron2026universat}
jointly accommodate spectral, spatial, and temporal variation within a single
architecture, making them the closest flexible EO baselines to our framework.

A separate line of work relaxes this requirement by changing the unit on which
the model operates. Presto~\citep{tseng2023lightweight} discards image
structure entirely, modeling individual pixel time series and relying on
temporal and multimodal signal instead, while
Atomizer~\citep{Turckheim_2025_BMVC} retains atomic measurements as the input
representation and processes them as an unordered set. More generally, Set
Transformer~\citep{lee2019set} showed that attention can operate directly on
unordered sets, while Perceiver~\citep{jaegle2021perceiver} introduced
cross-attention to a compact latent bottleneck and Perceiver
IO~\citep{jaegle2021perceiverio} extended this paradigm to query-defined
outputs.

\begin{figure}[t!]
  \centering
  \vspace{-0.25cm}
  \includegraphics[width=1.0\textwidth]{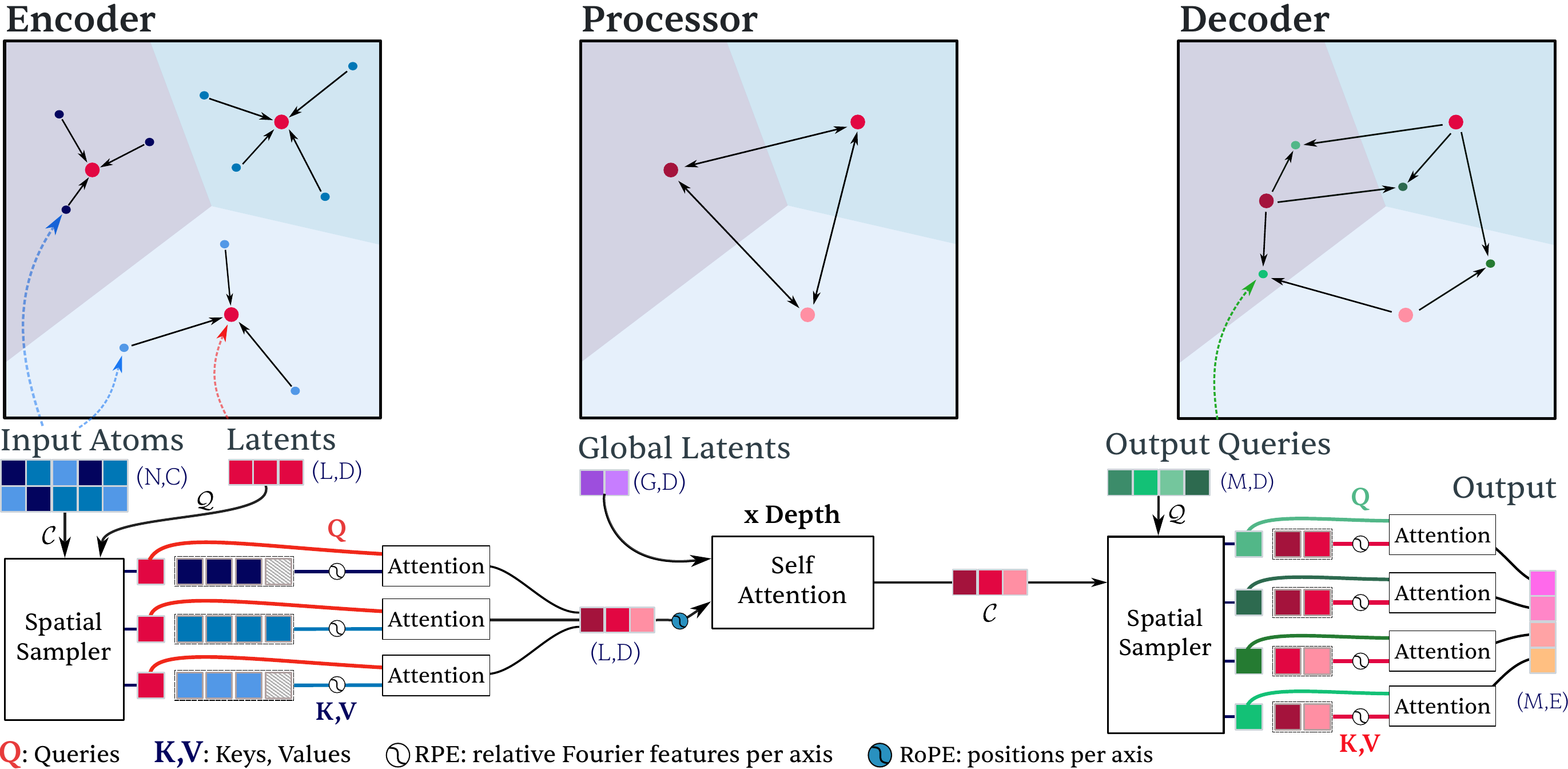}
  \vspace{-0.55cm}
  \caption{
The Atomizer-IO architecture. Atoms are locally aggregated into
a compact spatial representation, processed globally in latent space, and
decoded through spatial output queries. The same local context-to-query
operation is used for encoding and decoding, with spatial relationships defined
in physical coordinates rather than by a fixed input grid.
}
  \label{fig:Atomizer}
  \vspace{-0.45cm}
\end{figure}

Moving beyond grids does not mean abandoning spatial structure. 3D point-cloud methods routinely operate on unordered, sparsely distributed points that do not fit the dense tensor representations common in 2D vision~\citep{guo2019review3d}. Various alternatives have been proposed: PointNet~\citep{qi2017pointnet} operates on sets, PointNet++~\citep{qi2017pointnet++} and KPConv~\citep{thomas2019kpconv} on multi-scale Euclidean neighborhoods, and Superpoint Transformer~\citep{robert2023spt}, not unlike our work, on geometric graphs over 3D partitions. However, these approaches are not designed to jointly accommodate multi-spectral, multi-temporal, and multi-resolution data.
This creates a tension: grid-based models preserve useful spatial structure but require observations to conform to a fixed layout, while set-based models gain flexibility but provide no geometry by default.
Atomizer-IO resolves this by enabling reasoning over an arbitrary Voronoi tessellation of space.
Sensor-specific gridded structures are confined to the interfaces, where input observations are either read in, or output predictions are queried out. 
\vspace{-0.3cm}
\section{Methodology}
\vspace{-0.3cm}

\noindent

Atomizer-IO builds on Atomizer~\citep{Turckheim_2025_BMVC}, which decomposes an input into an unordered set of atoms.
Where Atomizer lets non-spatial latents cross-attend to all atoms, and only for image classification, Atomizer-IO introduces locality.
Each latent corresponds to a spatial anchor and attends only to nearby atoms, while still exchanging information with every other latent.
Spatially dense outputs are produced by querying the latents near any requested location.

This section follows one observation through the model (Figure~\ref{fig:Atomizer}): token construction (Section~\ref{sec:token_construction}), neighborhood assignment (Section~\ref{sec:spatial_sampler}), aggregation into a spatial latent (Section~\ref{sec:encoder}), latent interaction (Section~\ref{sec:processor}), and output queries (Section~\ref{sec:decoder}). Notation is summarized in Appendix~\ref{app:notation}.

\subsection{Token Construction}
\label{sec:token_construction}

\begin{figure}[ht]
  \centering
  \vspace{-0.25cm}
  \includegraphics[width=1.0\textwidth]{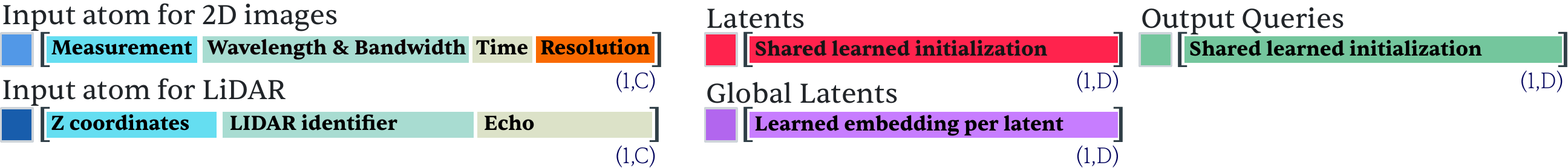}
  \vspace{-0.55cm}
  \caption{
Token construction in Atomizer-IO. 2D imagery and LiDAR share the same atomic
interface of measurements and acquisition metadata; spatial latents, global
latents, and output queries use learned initializations.
}
  \label{fig:encoder}
\vspace{-0.3cm}
\end{figure}

Atomizer-IO represents an input as a set of elementary observations, or
\emph{atoms}. Each atom holds a measured value $v \in \mathbb{R}$ (e.g. reflectance) and metadata
$u_1,\dots,u_J$ describing its acquisition. We combine these into tokens
$\mathbf{f} = [\phi_0(v), \phi_1(u_1), \dots, \phi_J(u_J)]$, where
$\phi_0,\dots,\phi_J$ are encoders for each metadata field (e.g. wavelength and resolution). The input is represented as
$\mathcal{F}=\{(\mathbf{f}_i,\mathbf{p}_i)\}_{i=1}^{N}$, pairing each token
with its spatial position $\mathbf{p}_i\in\mathbb{R}^2$.
Appendix~\ref{app:token_encoding} further specifies our encoding schemes.
To prevent the model from learning position-specific features, $\mathbf{p}_i$ is not included in $\mathbf{f}_i$ and is used only
by the spatial sampler (Section~\ref{sec:spatial_sampler}).
An atom's metadata fields follow from the sensor, and the architecture is
indifferent to which ones it receives (Figure~\ref{fig:encoder}). For 2D
rasters, $v$ is the value measured at one pixel, in one spectral band, at one
acquisition time, and the metadata describe that context: ground sampling
distance, spectral support, and acquisition date. For 3D point clouds, $v$ is
the elevation, with a sensor identifier, the position of the return within the
LiDAR pulse, and the return intensity when available as metadata. Nothing in
the representation depends on a channel count or a grid, and atoms of different
kinds can coexist in the same input set, as when LiDAR returns and co-registered
image pixels are processed together.

\subsection{Spatial Sampler}
\label{sec:spatial_sampler}

\begin{figure}[ht]
\centering
\vspace{-0.2cm}
\includegraphics[width=1.0\textwidth]{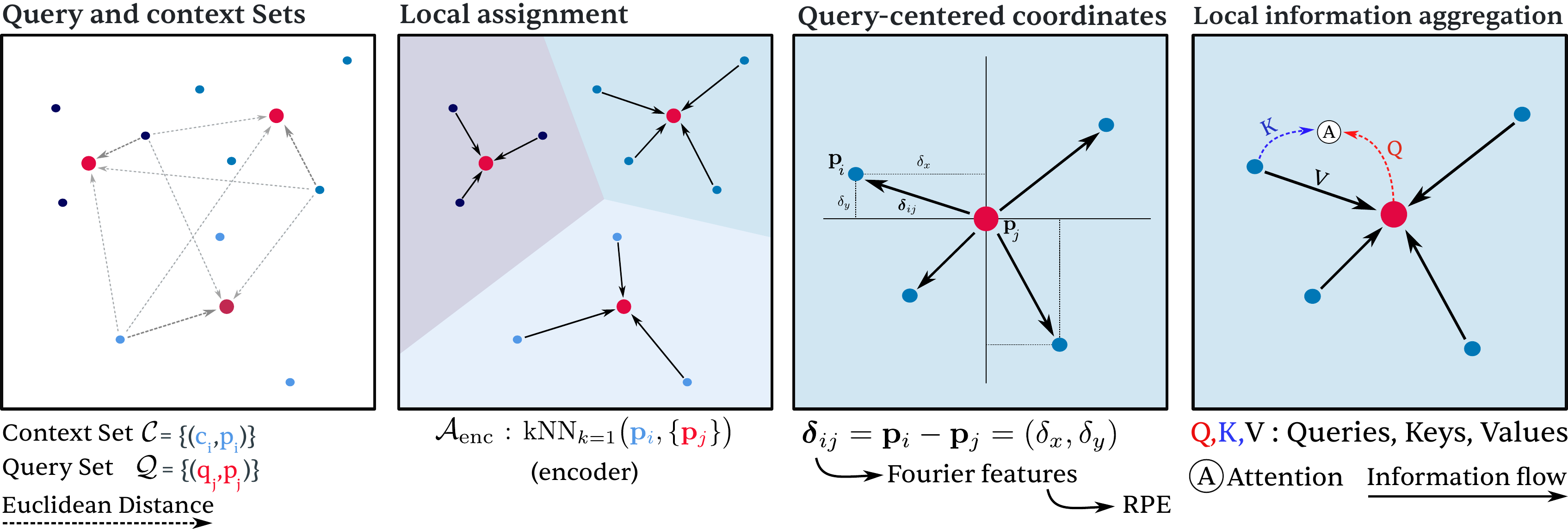}
\vspace{-0.5cm}
\caption{
Spatial sampler. From left to right: query and context sets; construction of a
local receptive field from spatial proximity, illustrated by the encoder's
nearest-latent assignment and resulting Voronoi partition; query-centered
relative coordinates; local cross-attention.
}
\label{fig:spatial_sampler}
\end{figure}

The spatial sampler (Figure~\ref{fig:spatial_sampler}) operates on two sets: a
\emph{query} set
$\mathcal{Q}=\{(q_j,\mathbf{p}_j)\}_{j=1}^{|\mathcal{Q}|}$ seeking information
from a \emph{context} set
$\mathcal{C}=\{(c_i,\mathbf{p}_i)\}_{i=1}^{|\mathcal{C}|}$.
It proceeds in two steps: first, it establishes which context elements belong
to the receptive field of each query; then, information flows from each
receptive field to its query. In the encoder, input atoms form the context set
and spatial latents the query set; in the decoder, spatial latents form the
context set and output locations the query set.

\vspace{-0.25cm}
\vspace{-0.25cm}

\paragraph{Assignment.}
We denote by
$\mathcal{A}(\mathcal{C},\mathcal{Q})
=\{\mathcal{V}_j\}_{j=1}^{|\mathcal{Q}|}$
the spatial assignment, where $\mathcal{V}_j$ is the index set of context
elements in the local receptive field of query $q_j$. Receptive fields are
built by nearest-neighbor search in the horizontal plane, which suits
bird's-eye-view reasoning in Earth observation and applies to both rasters and
3D point clouds. In the encoder, each context element is assigned to its nearest query, inducing a Voronoi partition, which we denote by
$\mathcal{A}_{\mathrm{enc}}$. In the decoder, each query instead gathers its
$k$ nearest context elements, which we denote by
$\mathcal{A}_{\mathrm{dec}}$. Implementation details are given in
Appendix~\ref{app:spatial_sampler_implementation}.

\vspace{-0.25cm}
\paragraph{Information flow.}
Cross-attention defines a one-way flow: each query gathers information from its
receptive field, while context elements themselves are not updated. Geometry is
therefore expressed from the perspective of the receiving query. For a context
element $i\in\mathcal{V}_j$, we use the relative offset
$\boldsymbol{\delta}_{ij}=\mathbf{p}_i-\mathbf{p}_j$ and encode it as
$\phi_\delta(\boldsymbol{\delta}_{ij})$ using Fourier features, analogous to the
positional encoding used in NeRF~\citep{mildenhall2021nerf}. The encoded offset
is concatenated to the context feature before information is passed to the
query through cross-attention
(Appendix~\ref{app:rpe_encoding}).

\vspace{-0.25cm}
\subsection{Encoder}
\label{sec:encoder}
\vspace{-0.15cm}

In the encoder, spatial latents form the query set and input atoms form the
context set of the spatial sampler.

\begin{wrapfigure}[14]{r}{0.49\textwidth}
    \centering
    \vspace{-0.95cm}
    \includegraphics[width=\linewidth]{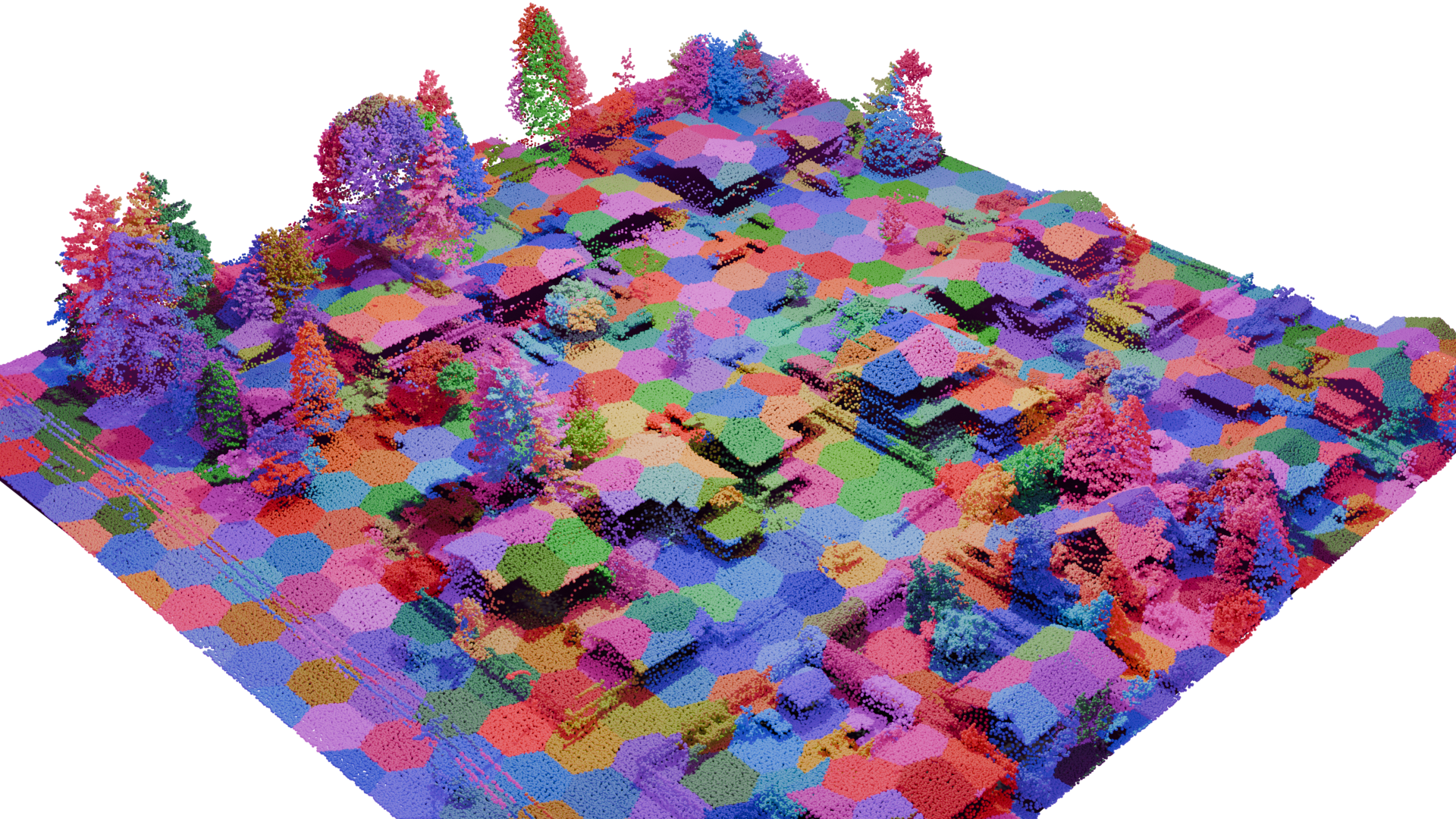}
    \caption{
    Example spatial latent partition on DALES
    (Section~\ref{sec:flexible_geometry}). Colors show the local receptive
    fields induced by assigning each input atom to its nearest spatial latent.
    }
    \label{fig:latent_partition}
    \vspace{-0.65cm}
\end{wrapfigure}
\vspace{-0.25cm}

\paragraph{Adaptive latent allocation.}
No operation in the architecture requires a fixed number or layout of spatial
latents. In practice, we scale the number $L$ of spatial latents with the number
of input observations and place them uniformly over the spatial extent of the input. The
values of $L$ and the layouts used in our experiments are given in
Appendix~\ref{app:training_details}. Each latent has an anchor position
$\mathbf{p}_\ell\in\mathbb{R}^2$ and shares the same learned initialization
$\mathbf{h}_{\mathrm{init}}\in\mathbb{R}^D$, where $D$ is the latent embedding
dimension. We denote the initialized latent set by
$\mathcal{H}^{\mathrm{init}}
=\{(\mathbf{h}_{\mathrm{init}},\mathbf{p}_\ell)\}_{\ell=1}^{L}$.
Figure~\ref{fig:latent_partition} illustrates the one instance of a spatial partition
on a DALES scene.

\vspace{-0.25cm}
\paragraph{Local encoding.}
For the encoder, the spatial sampler returns the assigned atoms for each anchored latent 
$\{\mathcal{V}_\ell\}_{\ell=1}^{L}
=\mathcal{A}_{\mathrm{enc}}
(\mathcal{F},\mathcal{H}^{\mathrm{init}})$.
For an atom $i\in\mathcal{V}_\ell$, we compute the relative offset
$\boldsymbol{\delta}_{i\ell}=\mathbf{p}_i-\mathbf{p}_\ell$, concatenate its
Fourier encoding to the atom feature, and process the result with a shared MLP:
$\mathbf{z}_{i\ell}=
\operatorname{MLP}([\mathbf{f}_i;
\phi_\delta(\boldsymbol{\delta}_{i\ell})])$.
Each spatial latent is then updated by local cross-attention over its processed
context,
$\mathbf{h}^{(0)}_\ell=
\operatorname{CrossAttn}_{\mathrm{enc}}(
\mathbf{h}_{\mathrm{init}},
\{\mathbf{z}_{i\ell}:i\in\mathcal{V}_\ell\})$.
The MLP and cross-attention weights are shared across latents, so the parameter
count is independent of $L$. The resulting encoded latent set is
$\mathcal{H}^{(0)}
=\{(\mathbf{h}^{(0)}_\ell,\mathbf{p}_\ell)\}_{\ell=1}^{L}$.

For dense local contexts, we randomly sample at most $m$ indices from each
$\mathcal{V}_\ell$ before cross-attention, bounding the attention cost per
latent. On Sen1Floods11 (Section~\ref{sec:eo_tasks}), where inputs contain 15
channels at $512\times512$ resolution, a global cross-attention mechanism would
need to discard $99.987\%$ of the input atoms to match the same number of
query--context interactions. Local sampling instead distributes this budget
across the spatial partition and exposes the model to different subsets of each
local context over training
(Appendix~\ref{app:spatial_sampler_complexity}).

\vspace{-0.25cm}
\subsection{Processor}
\label{sec:processor}
\vspace{-0.15cm}

The processor operates on the encoded spatial latent set
$\mathcal{H}^{(0)}
=\{(\mathbf{h}^{(0)}_\ell,\mathbf{p}_\ell)\}_{\ell=1}^{L}$
together with $G$ learned global latents
$\mathcal{G}^{(0)}=\{\mathbf{g}^{(0)}_r\}_{r=1}^{G}$, with
$\mathbf{g}^{(0)}_r\in\mathbb{R}^D$, similar in spirit to register
tokens~\citep{darcet2024vision}. A processor of depth $B$ stacks global
multi-head self-attention blocks over the spatial and global latents,
$(\mathcal{H}^{(b)},\mathcal{G}^{(b)})
=\operatorname{Block}_b(
\mathcal{H}^{(b-1)},\mathcal{G}^{(b-1)})$
for $b=1,\dots,B$, allowing information gathered from different local
neighborhoods to propagate across the latent representation. The final
processed spatial latent set is $\mathcal{H}^{(B)}$.

\vspace{-0.25cm}
\paragraph{Latent spatial processing.}
While the encoder introduces atom--latent geometry through relative positional
encoding in the spatial sampler, spatial relationships \emph{between latents}
are represented using two-dimensional rotary positional encoding
(RoPE)~\citep{heo2024rotary} based on their anchor positions
$\mathbf{p}_\ell$. Because the distance between anchors depends on how they are
distributed over the input extent, no single compression scale suits every
input; we therefore use a learned scale $s_p>0$, initialized to a reference scale
$s_0$ (Appendix~\ref{app:compression}). Coordinates are compressed with $s_p$ before applying RoPE to the queries and keys of the spatial latents. Global latents
have no associated spatial position and are therefore left unrotated.

\vspace{-0.25cm}
\paragraph{Temporal processing.}
Multitemporal inputs require no dedicated temporal module: atoms from all
acquisitions can be encoded jointly in the same input set, with acquisition
time provided as metadata. When explicit temporal aggregation is preferred, the
encoder instead processes each timestep independently, with shared weights and
identical spatial anchor positions, producing
$\mathcal{H}^{(0)}_t
=\{(\mathbf{h}^{(0)}_{t\ell},\mathbf{p}_\ell)\}_{\ell=1}^{L}$
for each timestep $t$. The processor is then applied independently within each
timestep, yielding
$\mathcal{H}^{(B)}_t
=\{(\mathbf{h}^{(B)}_{t\ell},\mathbf{p}_\ell)\}_{\ell=1}^{L}$.
For each spatial anchor $\ell$, the $T$ representations
$\{\mathbf{h}^{(B)}_{t\ell}\}_{t=1}^{T}$ are aggregated along the temporal axis
by a small transformer with a learned aggregation token, using temporal RoPE
with the same coordinate compression and a dedicated learned scale. This
produces a temporally aggregated representation
$\bar{\mathbf{h}}_\ell$ for each anchor, defining
$\bar{\mathcal{H}}^{(B)}
=\{(\bar{\mathbf{h}}_\ell,\mathbf{p}_\ell)\}_{\ell=1}^{L}$.
When temporal aggregation is used, this set replaces $\mathcal{H}^{(B)}$ as
decoder context. Its contribution is evaluated in
Appendix~\ref{app:temporal_ablation}.

\vspace{-0.25cm}
\subsection{Decoder}
\label{sec:decoder}
\vspace{-0.15cm}

The decoder predicts at a set of requested spatial positions
$\{\mathbf{p}_j\}_{j=1}^{M}$. Each output query shares the same learned
initialization $\mathbf{q}_{\mathrm{init}}\in\mathbb{R}^D$, giving the
initialized query set
$\mathcal{Q}^{\mathrm{init}}
=\{(\mathbf{q}_{\mathrm{init}},\mathbf{p}_j)\}_{j=1}^{M}$.
Below, $\mathcal{H}^{(B)}$ denotes the decoder context, replaced by
$\bar{\mathcal{H}}^{(B)}$ when explicit temporal aggregation is used. The
spatial sampler returns
$\{\mathcal{V}_j\}_{j=1}^{M}
=\mathcal{A}_{\mathrm{dec}}
(\mathcal{H}^{(B)},\mathcal{Q}^{\mathrm{init}})$.
For a latent $\ell\in\mathcal{V}_j$, we compute the relative offset
$\boldsymbol{\delta}_{\ell j}=\mathbf{p}_\ell-\mathbf{p}_j$
and concatenate its Fourier encoding to the latent feature. Each output query
is then obtained by local cross-attention over this context,
$\mathbf{q}_j=
\operatorname{CrossAttn}_{\mathrm{dec}}
(\mathbf{q}_{\mathrm{init}},
\{[\mathbf{h}^{(B)}_\ell;
\phi_\delta(\boldsymbol{\delta}_{\ell j})]:
\ell\in\mathcal{V}_j\})$,
and passed to a task-specific prediction head; for dense prediction, the input
$\mathbf{q}_{\mathrm{init}}$ is replaced by the input-conditioned query
$\tilde{\mathbf{q}}_j$ defined below. Because output locations are queried
explicitly, the decoder does not need to evaluate every spatial position
uniformly. We exploit this property in
Section~\ref{sec:flexible_decoding}.

For dense prediction, we additionally condition each output query on the
observations at its target location before it attends to the latent
representation, a local pathway analogous to the skip connections of
U-Net~\citep{ronneberger2015u}. Let
$\mathcal{S}_j\subseteq\{1,\dots,N\}$ be the indices of atoms co-located with
query $j$: the atoms at the pixel being predicted for a 2D raster, and the
atoms at the queried point for a point cloud. The query is first updated by
cross-attention over their features,
$\tilde{\mathbf{q}}_j=
\operatorname{CrossAttn}_{\mathrm{obs}}
(\mathbf{q}_{\mathrm{init}},
\{\mathbf{f}_i:i\in\mathcal{S}_j\})$,
combining observations at the target location with the processed information
its surroundings provide through the latent cross-attention above.
\vspace{-0.25cm}
\section{Experiments}
\label{sec:experiments}
\vspace{-0.15cm}
We evaluate Atomizer-IO by progressively relaxing the assumptions of a regular
grid. We begin with conventional raster Earth observation tasks, then vary input
composition and output density, remove the input grid entirely with irregular
3D point clouds, and finally test spatial reasoning under unseen input
geometries.
\vspace{-0.1cm}
\subsection{Performance Across Diverse EO Tasks}
\vspace{-0.1cm}
\label{sec:eo_tasks}

Architectures that relax fixed input structure for greater generality may give
up useful task- or sensor-specific priors, requiring more of the task structure
to be learned from data. Before relaxing the grid, we therefore verify that
Atomizer-IO is competitive when it is fully present, across a deliberately
heterogeneous suite of eight datasets spanning different sensing modalities,
channel counts, temporal lengths, spatial resolutions, input sizes, and
prediction problems (Table~\ref{tab:tasks}). Each dataset is treated as an
independent task and a separate model is trained from scratch.

\begin{table*}[ht]

\centering
\caption{
Datasets used for evaluation, spanning raster Earth observation and irregular
LiDAR inputs. \textbf{T}: timesteps; \textbf{C}: input channels;
\textbf{Geometry}: spatial structure of the input; \textbf{GSD}: ground
sampling distance, undefined for point clouds; \textbf{Out. C}: output classes
or regression targets.
}
\vspace{-0.25cm}
\label{tab:tasks}

\footnotesize
\setlength{\tabcolsep}{3pt}
\renewcommand{\arraystretch}{1.08}

\begin{tabular}{@{}llccllcc@{}}
\toprule
\textbf{Dataset} & \textbf{Sensor} & \textbf{T} & \textbf{C}
& \textbf{Geometry} & \textbf{GSD} & \textbf{Problem} & \textbf{Out. C} \\
\midrule
ForestNet~\citep{irvin2020forestnet} 
& Landsat-8 & 1 & 6 & $332^2$ & 15\,m & cls & 12 \\

BurnScars~\citep{jakubik2023foundation} 
& HLS & 1 & 6 & $512^2$ & 30\,m & seg & 2 \\

EuroSAT~\citep{helber2019eurosat} 
& Sentinel-2 & 1 & 13 & $64^2$ & 10\,m & cls & 10 \\

Cashew~\citep{jin2021smallholder} 
& Sentinel-2 & 1 & 13 & $256^2$ & 10\,m & seg & 7 \\

Sen1Floods11~\citep{rambour2020flood} 
& S1 + S2 & 1 & 15 & $512^2$ & 10\,m & seg & 2 \\

xView2~\citep{gupta2019xbd} 
& RGB (VHR) & 2 & 3 & $512^2$ & 0.5\,m & seg & 5 \\

BioMassters~\citep{nascetti2023biomassters} 
& S1 + S2 & 3 & 14 & $256^2$ & 10\,m & reg & 1 \\

PASTIS~\citep{garnot2021panoptic} 
& S1 + S2 & 6 & 10 & $128^2$ & 10\,m & seg & 19 \\

\midrule
DALES~\citep{varney2020dales} 
& LiDAR & 1 & 1 & point set & -- & 3D seg & 8 \\

FRACTAL~\citep{gaydon2024fractal} 
& LiDAR + VHR & 1 & 4 & points + image & -- & 3D seg & 7 \\

\bottomrule
\end{tabular}
\vspace{-0.25cm}
\end{table*}
We compare Atomizer-IO with grid-based baselines (ResNet50 ~\citep{he2016deep} and ViT ),
flexible EO-specific models (RAMEN and UniverSat), and Perceiver-IO as a generic
set-based latent architecture. All models are trained from scratch under
matched conditions (Appendix \ref{app:training_details}).

\begin{table}[t]
\centering
\caption{Single-task performance. Best in \textbf{bold}, second best
\underline{underlined}. Classification uses macro F1, segmentation mIoU,
and BioMassters RMSE (Mg/ha, $\downarrow$).}

\label{tab:single_task_results}
\footnotesize
\setlength{\tabcolsep}{3pt}
\renewcommand{\arraystretch}{1.05}
\begin{tabular}{@{}lcccccccc@{}}
\toprule
\textbf{Model}
& EuroSAT
& ForestNet
& BurnScars
& Sen1Floods
& PASTIS
& xView2
& Cashew
& BioMassters $\downarrow$ \\
\midrule
ResNet50
& \underline{90.9} & \underline{43.8} & 82.8 & 87.8
& 30.0 & 53.8 & \underline{69.9} & 46.3 \\

ViT
& 90.7 & 39.1 & 87.5 & 85.0
& 35.1 & 52.4 & 47.5 & 52.7 \\

RAMEN
& \textbf{92.1} & 40.4 & \underline{88.2} & 89.1
& \textbf{45.5} & 53.3 & 64.2 & \underline{44.7} \\

UniverSat
& 90.8 & \textbf{44.3} & 88.1 & 86.8
& 42.9 & \textbf{56.3} & 62.4 & 46.0 \\

Perceiver-IO
& \underline{90.9} & 35.4 & 83.4 & \underline{92.2}
& 17.9 & 43.8 & 26.2 & 52.2 \\

\textbf{Atomizer-IO}
& 90.2 & 35.8 & \textbf{88.8} & \textbf{93.2}
& \underline{44.1} & \underline{54.6} & \textbf{72.0} & \textbf{42.4} \\
\bottomrule
\end{tabular}
\vspace{-0.5cm}
\end{table}
Across the eight tasks, Atomizer-IO excels at spatially-explicit tasks, ranking first on four and second on two (Table~\ref{tab:single_task_results}), matching or exceeding grid-based and flexible EO architectures on their own terms, on raster data. This establishes the baseline for the relaxations that follow.
These tasks rely on different sources of information. PASTIS benefits strongly from reasoning across acquisitions, Sen1Floods11 from local spectral and radar measurements, while Cashew shows a stronger dependence on
spatial context. Appendix~\ref{app:ablations} examines these differences directly by ablating temporal aggregation, spatial reasoning, and local observation conditioning across representative tasks. As expected, we also observe that Atomizer-IO does not have an edge over Perceiver-IO for non spatially explicit classification tasks, such as EuroSAT and ForestNet.

\subsection{Flexible Input Composition: Handling Missing Channels}
\label{sec:flexible_input}

In Earth observation, available channels can change with sensor availability, weather conditions or spectral coverage.
Models built around a fixed channel layout must preserve that layout when observations are missing, whereas flexible representations can omit unavailable observations directly.
We evaluate this setting on Sen1Floods11 and BioMassters, both combining Sentinel-1 and Sentinel-2 observations. 
In this experiment, all models use the same band-dropout augmentation, so variation in channel availability is anticipated during
training. Atomizer-IO and UniverSat omit unavailable observations, whereas fixed-channel models are zero-padded at the corresponding positions. The same trained model is evaluated under six input configurations, from the full
observation set to single-sensor and restricted spectral subsets.

\vspace{-0.25cm}
\begin{table}[H]
\centering
\caption{
Performance under flexible input composition on Sen1Floods11
(mIoU, \%) and BioMassters (RMSE, Mg/ha).
All models are trained under the same band-dropout protocol.
}
\label{tab:modality_drop}
\footnotesize
\setlength{\tabcolsep}{3pt}
\renewcommand{\arraystretch}{1.05}
\vspace{-0.15cm}
\begin{tabular}{@{}lcccccccccccc@{}}
\toprule
& \multicolumn{6}{c}{\textbf{Sen1Floods11} $\uparrow$}
& \multicolumn{6}{c}{\textbf{BioMassters} $\downarrow$} \\
\cmidrule(lr){2-7} \cmidrule(l){8-13}
\textbf{Test Configuration}
& \textbf{Atom.}
& \textbf{Res.}
& \textbf{ViT}
& \textbf{Perc.}
& \textbf{RAM.}
& \textbf{Uni.}
& \textbf{Atom.}
& \textbf{Res.}
& \textbf{ViT}
& \textbf{Perc.}
& \textbf{RAM.}
& \textbf{Uni.} \\
\midrule

All bands
& \textbf{92.9} & 88.1 & 84.5 & \underline{92.1} & 87.0 & 88.7
& \textbf{41.5} & 47.6 & 52.9 & 52.0 & \underline{43.4} & 45.7 \\

S2 only
& \textbf{92.8} & 86.4 & 84.4 & \underline{92.2} & 87.8 & 88.6
& \textbf{45.8} & 55.9 & 55.5 & 57.7 & \underline{48.4} & 49.1 \\

S1 only
& \textbf{80.0} & 66.4 & 75.3 & 77.2 & 75.1 & \underline{79.4}
& \textbf{48.1} & 52.1 & 55.5 & 58.6 & 49.2 & \underline{49.1} \\

RGB only
& \textbf{44.8} & 43.7 & 43.8 & 43.8 & \underline{44.5} & 43.8
& 71.0 & 73.8 & 71.4 & \textbf{70.4} & 71.9 & \underline{70.8} \\

No SWIR
& \textbf{87.8} & 49.0 & 75.1 & 81.8 & 63.7 & \underline{85.2}
& \textbf{43.0} & 49.1 & 54.0 & 53.4 & \underline{46.0} & 46.7 \\

No red-edge
& \textbf{92.9} & 85.3 & 80.5 & \underline{89.1} & 85.9 & 88.0
& \textbf{42.2} & 49.8 & 53.8 & 54.0 & \underline{45.4} & 46.4 \\

\midrule
\textbf{Average}
& \textbf{81.9} & 69.8 & 73.9 & \underline{79.4} & 74.0 & 78.9
& \textbf{48.6} & 54.7 & 57.2 & 57.7 & \underline{50.7} & 51.3 \\

\bottomrule
\end{tabular}

\vspace{2pt}
{\scriptsize
Atom.: Atomizer-IO; Res.: ResNet50; Perc.: Perceiver-IO;
RAM.: RAMEN; Uni.: UniverSat.
}
\end{table}
\vspace{-0.4cm}
Table~\ref{tab:modality_drop} shows that Atomizer-IO achieves the best average
performance on both datasets and leads in most configurations. Its advantage is
largest when specific bands or an entire sensor are removed, but nearly
disappears under RGB only, where much of the discriminative information is
lost. More broadly, architectures that relax the fixed channel layout tend to
handle changes in input composition better than conventional fixed-layout
baselines, supporting the value of decoupling representation from a prescribed
channel structure.
\vspace{-0.25cm}
\subsection{Flexible Output Decoding: Inference-Time Compute Control}
\label{sec:flexible_decoding}
\vspace{-0.1cm}

Here we study various relaxations of the output grid. Atomizer-IO predicts at explicitly requested spatial locations, so dense segmentation corresponds simply to placing one
query at every pixel (\emph{Dense} decoding). Because the decoder does not require
all locations to be queried uniformly, output density can instead be adapted to
the spatial complexity of the prediction.

We exercise this property by approximating dense predictions while querying
fewer locations. For each region, we first probe a small number of locations.
If all probes predict the same class, that class is assigned to the remaining
locations without querying them; otherwise, additional computation is allocated
to the region. \emph{Zone-probe} applies this rule to regions associated with
individual latents, while \emph{Quadtree} recursively subdivides disagreeing
regions and probes them at finer spatial scales. All strategies use the same
trained checkpoint and differ only in how densely the output space is queried.
Details and complexity are given in Appendix~\ref{app:decode_strategies}.\begin{figure}[t!]
  \centering
  \vspace{-0.15cm}
  \includegraphics[width=1.0\textwidth]{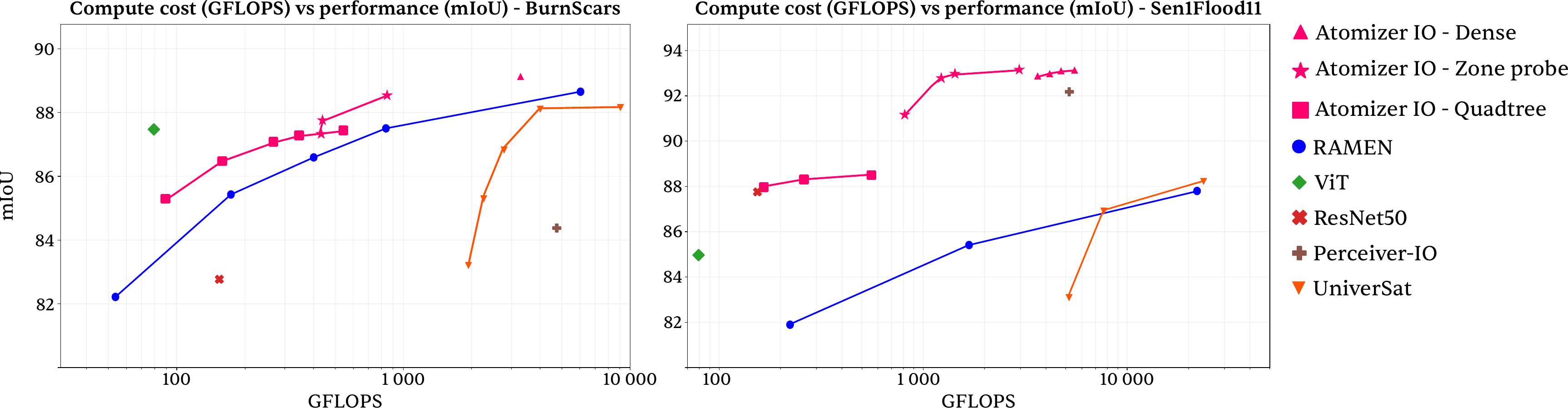}
  \vspace{-0.45cm}
  \caption{
Compute cost (GFLOPs) versus segmentation performance (mIoU). Dense,
Zone-probe, and Quadtree decoding vary inference cost through different output
query strategies, using the same Atomizer-IO checkpoint without retraining.
}
  \label{fig:compute_tradeoff}
  \vspace{-0.45cm}
\end{figure}
Figure~\ref{fig:compute_tradeoff} shows that a single trained checkpoint spans
a broad range of inference costs, with \emph{Dense} decoding being
competitive with the flexible EO baselines at the high-compute end. The two adaptive strategies
reduce cost differently: \emph{Zone-probe} couples output regions to the
model's latent partition, while \emph{Quadtree} adapts output density
independently of that partition through recursive refinement.

Atomizer-IO therefore exposes complementary inference-time controls: like
RAMEN and UniverSat, compute can be reduced through a coarser latent
representation, while the query-based decoder additionally allows output
density to vary after training. Probe agreement is a heuristic rather than a
guarantee, reflecting the trade-off between exact dense evaluation and reduced
inference cost.\begin{wraptable}[21]{r}{0.48\textwidth}
\centering
\caption{DALES and FRACTAL segmentation results (mIoU, \%).}
\label{tab:lidar_results}
\vspace{-0.45cm}
\vspace{-\baselineskip}
\setlength{\tabcolsep}{5pt}
\renewcommand{\arraystretch}{1.03}
\vspace{0.7cm}
\begin{tabular}{@{}lc@{}}

\multicolumn{2}{c}{\textbf{DALES}} \\
\toprule
\textbf{Method} & \textbf{mIoU} \\
\midrule
KPConv~\citep{thomas2019kpconv}                  & \textbf{81.1} \\
SPT~\citep{robert2023spt}                        & 79.6 \\
PointNet++~\citep{qi2017pointnet++}              & 68.3 \\
ConvPoint~\citep{boulch2020convpoint}            & 67.4 \\
\textbf{Atomizer-IO}                             & 66.7 \\
SPG~\citep{landrieu2018spg}             & 60.6 \\
PointCNN~\citep{li2018pointcnn}                  & 58.4 \\
ShellNet~\citep{zhang2019shellnet}               & 57.4 \\
Perceiver-IO~\citep{jaegle2021perceiverio}       & 48.7 \\
\bottomrule

\addlinespace[2pt]

\multicolumn{2}{c}{\textbf{FRACTAL}} \\
\toprule
\textbf{Atomizer-IO}                             & \textbf{78.4} \\
Perceiver-IO~\citep{jaegle2021perceiverio}     & 78.1 \\
RandLA-Net~\citep{hu2020randla}                & 77.5 \\

\bottomrule

\end{tabular}

\end{wraptable}
\vspace{-0.65cm}

\subsection{Flexible Input Geometry:\\Unordered LiDAR Point Clouds}
\label{sec:flexible_geometry}


We break free of the input grid for semantic segmentation of 3D point clouds.
Our goal is not to outperform 3D-specific architectures, but to demonstrate the
flexibility of our framework. On DALES and FRACTAL
(Table~\ref{tab:tasks}), each LiDAR point becomes an atom
(Appendix~\ref{app:echo-encoding}). On FRACTAL, RGB pixels from co-registered
orthophotos are additionally encoded as atoms, allowing both modalities to be
jointly processed. Training and inference tiling, point sampling, and batching
details are given in Appendix~\ref{app:lidar_protocol}.

Atomizer-IO is competitive on both datasets without 3D-specific design beyond its tokenizer, ahead of three of the DALES 3D baselines and best on FRACTAL (Table~\ref{tab:lidar_results}).
Perceiver-IO performs poorly on DALES, highlighting the importance of geometric structure in the latent reasoning, but both models outperform RandLA-Net on FRACTAL, suggesting the image modality alone may already be a strong predictor for this benchmark.
Our gap to KPConv and SPT on DALES suggests opportunities for geometry-adaptive 3D latent partitions, hierarchical reasoning, and finer local 3D modeling. We leave these extensions to future work.

\par
\FloatBarrier
\par
\vspace{-0.3cm}
\subsection{Controlled Analysis of Spatial Structure}
\label{sec:controlled_analysis}

\begin{wraptable}[9]{l}{0.36\textwidth}
\vspace{-0.05cm}
\centering
\caption{
RoPE ablation. Accuracy on MNIST and mIoU on Sen1Floods11.
}
\setlength{\tabcolsep}{6pt}
\renewcommand{\arraystretch}{1.05}
\begin{tabular}{lcc}
\toprule
& \textbf{MNIST} & \textbf{Sen1Fl11} \\
\midrule
No RoPE & 29.48 & 92.42 \\
RoPE    & \textbf{99.37} & \textbf{93.17} \\
\bottomrule
\end{tabular}

\label{tab:rope_ablation}
\vspace{-0.3cm}
\end{wraptable}

Removing the input grid does not remove the need for spatial structure. In
remote-sensing tasks, however, this is difficult to isolate because individual
observations can already be highly informative: a local spectral signature may
indicate water, vegetation, or other materials without requiring much spatial
context. We therefore use MNIST as a controlled setting. Because MNIST images
are nearly binary, individual pixel intensities carry little information in
isolation, and classification depends primarily on their spatial arrangement.

Atomizer-IO's spatial latents share the same learned initialization and
attention weights and are distinguished by the observations they aggregate and
their relative geometry. On MNIST, each latent carries only local intensity
information; removing RoPE therefore prevents the processor from recovering
their spatial arrangement and reduces accuracy from $99.37$ to $29.48$. On
Sen1Floods11, in contrast, removing RoPE only reduces mIoU from $93.17$ to
$92.42$, consistent with the strong predictive content already present in the
local multispectral and radar observations. Broader ablations in
Appendix~\ref{app:ablations} further examine how different tasks rely on local
observation content, spatial reasoning, and temporal structure.

\begin{wraptable}[13]{r}{0.41\textwidth}
\vspace{-0.5cm}
\centering
\caption{
MNIST accuracy (\%) under unseen input geometries. Digit pixels are always
retained; both models use the same checkpoint trained on complete inputs.
}
\label{tab:mnist_token_drop_darkonly}
\setlength{\tabcolsep}{6pt}
\renewcommand{\arraystretch}{0.9}
\begin{tabular}{lcc}
\toprule
\thead[l]{\textbf{Background}\\\textbf{kept} (\%)} & \textbf{Atomizer-IO} & \textbf{ViT} \\
\midrule
$100$ & \textbf{99.26} & 98.11 \\
$75$ & \textbf{98.72} & 97.65 \\
$50$ & \textbf{98.37} & 94.85 \\
$25$ & \textbf{94.93} & 72.17 \\
$10$ & \textbf{74.36} & 37.24 \\
\bottomrule
\end{tabular}
\vspace{-0.3cm}
\end{wraptable}

We finally test whether the same model can process input geometries not seen
during training. Atomizer-IO and a parameter-matched ViT with $1\times1$ pixel
patches are trained only on complete MNIST images. At test time, the same
background pixels are removed entirely from both models while all digit pixels
are retained, so both receive exactly the same surviving observations. As the
background becomes increasingly sparse and irregular, Atomizer-IO degrades more
gracefully, retaining $94.93\%$ accuracy when only $25\%$ of the background
pixels are kept, compared with $72.17\%$ for ViT
(Table~\ref{tab:mnist_token_drop_darkonly}). This suggests that Atomizer-IO
better preserves its ability to exploit task-relevant observations under
substantial changes in input geometry.
\vspace{-0.25cm}
\section{Conclusion}
\vspace{-0.25cm}

Computing on a regular grid is not required for competitive performance under
a competitive compute budget. Across eight Earth-observation benchmarks and two
LiDAR datasets spanning classification, regression, and 2D/3D segmentation,
Atomizer-IO remains competitive with grid-based and flexible EO architectures
while progressively relaxing assumptions on input and output structure. To our
knowledge, no other EO architecture has been evaluated on both multispectral
rasters and unordered LiDAR point clouds through the same encoder and without
modality-specific processing branches.

Unlike architectures whose structure reflects assumptions about the data,
Atomizer-IO encodes sensing information explicitly in its input tokens. Each
atom represents a measurement at a particular place, time, and sensor, without requiring a fixed shape, channel count, or neighboring measurement; locality and geometry are introduced afterwards from physical relationships between observations.

As future work, we expect this perspective to be particularly relevant to large-scale pretraining, where a central challenge is reconciling observations that differ in what they measure and how they are sampled. Existing architectures typically begin from a tensor representation and introduce dedicated mechanisms to accommodate each axis of variation. Atomizer-IO starts from the opposite premise: the observation itself is the common unit. The results in this paper, in which the exact same architecture provides competitive results across a wide range of modalities, suggest that Atomizer-IO could make a good candidate architecture for a multi-modal EO foundation model.

Our results allow us to conclude that pixels, patches, and grids need not define the interface of a sensing architecture. They can instead be treated as particular discretizations of observations, introduced when useful rather than assumed from the start.

\clearpage

\subsection*{AI use statement}

In this work, we used generative AI tools to implement the methods, correct grammar, and edit the research paper to improve readability. We have not used generative AI tools to generate synthetic data sets, develop theoretical models, formulate or prove mathematical claims, propose hypotheses, design experiments, translate text, clean data sets, or interpret results, and all other required disclosure tasks are not applicable to this work. We have reviewed all AI-assisted work. Specifically, all AI-generated code and text suggestions were thoroughly reviewed, manually verified for correctness, and substantially modified by the authors. We take responsibility for the final content of this work, including text, claims, or artifacts produced with the aid of generative AI.
\clearpage

\bibliographystyle{iclr2027_conference}
\bibliography{references}

@String(CVPR  = {IEEE Conf. Comput. Vis. Pattern Recog.})

@String(ICCV  = {Int. Conf. Comput. Vis.})

@String(BMVC  = {Brit. Mach. Vis. Conf.})

@String(CVPRW = {IEEE Conf. Comput. Vis. Pattern Recog. Worksh.})

@String(CVPR  = {CVPR})

@String(ICCV  = {ICCV})

@String(BMVC  =	{BMVC})

@String(CVPRW = {CVPRW})

@article{dosovitskiy2020image,
  title={An image is worth 16x16 words: Transformers for image recognition at scale},
  author={Dosovitskiy, Alexey and Beyer, Lucas and Kolesnikov, Alexander and Weissenborn, Dirk and Zhai, Xiaohua and Unterthiner, Thomas and Dehghani, Mostafa and Minderer, Matthias and Heigold, Georg and Gelly, Sylvain and others},
  journal={arXiv preprint arXiv:2010.11929},
  year={2020}
}

@inproceedings{he2016deep,
  title={Deep residual learning for image recognition},
  author={He, Kaiming and Zhang, Xiangyu and Ren, Shaoqing and Sun, Jian},
  booktitle={Proceedings of the IEEE conference on computer vision and pattern recognition},
  pages={770--778},
  year={2016}
}

@inproceedings{ronneberger2015u,
  title={U-net: Convolutional networks for biomedical image segmentation},
  author={Ronneberger, Olaf and Fischer, Philipp and Brox, Thomas},
  booktitle={International Conference on Medical image computing and computer-assisted intervention},
  pages={234--241},
  year={2015},
  organization={Springer}
}

@article{weinstein2021benchmark,
  title={A benchmark dataset for canopy crown detection and delineation in co-registered airborne RGB, LiDAR and hyperspectral imagery from the National Ecological Observation Network},
  author={Weinstein, Ben G and Graves, Sarah J and Marconi, Sergio and Singh, Aditya and Zare, Alina and Stewart, Dylan and Bohlman, Stephanie A and White, Ethan P},
  journal={PLoS computational biology},
  volume={17},
  number={7},
  pages={e1009180},
  year={2021},
  publisher={Public Library of Science San Francisco, CA USA}
}

@article{mildenhall2021nerf,
  title={Nerf: Representing scenes as neural radiance fields for view synthesis},
  author={Mildenhall, Ben and Srinivasan, Pratul P and Tancik, Matthew and Barron, Jonathan T and Ramamoorthi, Ravi and Ng, Ren},
  journal={Communications of the ACM},
  volume={65},
  number={1},
  pages={99--106},
  year={2021},
  publisher={ACM New York, NY, USA}
}

@article{li2018pointcnn,
  title={Pointcnn: Convolution on x-transformed points},
  author={Li, Yangyan and Bu, Rui and Sun, Mingchao and Wu, Wei and Di, Xinhan and Chen, Baoquan},
  journal={Advances in neural information processing systems},
  volume={31},
  year={2018}
}

@inproceedings{qi2017pointnet,
  title={Pointnet: Deep learning on point sets for 3d classification and segmentation},
  author={Qi, Charles R and Su, Hao and Mo, Kaichun and Guibas, Leonidas J},
  booktitle={Proceedings of the IEEE conference on computer vision and pattern recognition},
  year={2017}
}

@article{qi2017pointnet++,
  title={Pointnet++: Deep hierarchical feature learning on point sets in a metric space},
  author={Qi, Charles Ruizhongtai and Yi, Li and Su, Hao and Guibas, Leonidas J},
  journal={Advances in neural information processing systems},
  volume={30},
  year={2017}
}

@article{boulch2020convpoint,
  title={ConvPoint: Continuous convolutions for point cloud processing},
  author={Boulch, Alexandre},
  journal={Computers \& Graphics},
  volume={88},
  pages={24--34},
  year={2020},
  publisher={Elsevier}
}

@inproceedings{landrieu2018spg,
  title={Large-scale point cloud semantic segmentation with superpoint graphs},
  author={Landrieu, Loic and Simonovsky, Martin},
  booktitle={2018 IEEE/CVF Conference on Computer Vision and Pattern Recognition},
  year={2018},
}

@inproceedings{robert2023spt,
  title={Efficient 3d semantic segmentation with superpoint transformer},
  author={Robert, Damien and Raguet, Hugo and Landrieu, Loic},
  booktitle={2023 IEEE/CVF International Conference on Computer Vision (ICCV)},
  year={2023},
}

@inproceedings{varney2020dales,
  title={DALES: A large-scale aerial LiDAR data set for semantic segmentation},
  author={Varney, Nina and Asari, Vijayan K and Graehling, Quinn},
  booktitle={2020 IEEE/CVF Conference on Computer Vision and Pattern Recognition Workshops (CVPRW)},
  pages={717--726},
  year={2020},
  organization={IEEE}
}

@inproceedings{zhang2019shellnet,
  title={Shellnet: Efficient point cloud convolutional neural networks using concentric shells statistics},
  author={Zhang, Zhiyuan and Hua, Binh-Son and Yeung, Sai-Kit},
  booktitle={2019 IEEE/CVF International Conference on Computer Vision (ICCV)},
  pages={1607--1616},
  year={2019},
  organization={IEEE}
}

@inproceedings{hu2020randla,
  title={Randla-net: Efficient semantic segmentation of large-scale point clouds},
  author={Hu, Qingyong and Yang, Bo and Xie, Linhai and Rosa, Stefano and Guo, Yulan and Wang, Zhihua and Trigoni, Niki and Markham, Andrew},
  booktitle={2020 IEEE/CVF Conference on Computer Vision and Pattern Recognition (CVPR)},
  pages={11105--11114},
  year={2020},
  organization={IEEE}
}

@article{nascetti2023biomassters,
  title={Biomassters: A benchmark dataset for forest biomass estimation using multi-modal satellite time-series},
  author={Nascetti, Andrea and Yadav, Ritu and Brodt, Kirill and Qu, Qixun and Fan, Hongwei and Shendryk, Yuri and Shah, Isha and Chung, Christine},
  journal={Advances in Neural Information Processing Systems},
  volume={36},
  pages={20409--20420},
  year={2023}
}

@article{gaydon2024fractal,
  title={Fractal: An ultra-large-scale aerial lidar dataset for 3d semantic segmentation of diverse landscapes},
  author={Gaydon, Charles and Daab, Michel and Roche, Floryne},
  journal={arXiv preprint arXiv:2405.04634},
  year={2024}
}

@inproceedings{garnot2021panoptic,
  title={Panoptic segmentation of satellite image time series with convolutional temporal attention networks},
  author={Garnot, Vivien Sainte Fare and Landrieu, Loic},
  booktitle={2021 IEEE/CVF International Conference on Computer Vision (ICCV)},
  pages={4852--4861},
  year={2021},
  organization={IEEE}
}

@article{gupta2019xbd,
  title={xbd: A dataset for assessing building damage from satellite imagery},
  author={Gupta, Ritwik and Hosfelt, Richard and Sajeev, Sandra and Patel, Nirav and Goodman, Bryce and Doshi, Jigar and Heim, Eric and Choset, Howie and Gaston, Matthew},
  journal={arXiv preprint arXiv:1911.09296},
  year={2019}
}

@article{rambour2020flood,
  title={Flood detection in time series of optical and sar images},
  author={Rambour, Cl{\'e}ment and Audebert, Nicolas and Koeniguer, E and Le Saux, Bertrand and Crucianu, Michel and Datcu, Mihai},
  journal={The International Archives of the Photogrammetry, Remote Sensing and Spatial Information Sciences},
  volume={43},
  number={B2},
  pages={1343--1346},
  year={2020}
}

@inproceedings{darcet2024vision,
  title={Vision transformers need registers},
  author={Darcet, Timoth{\'e}e and Oquab, Maxime and Mairal, Julien and Bojanowski, Piotr},
  booktitle={International conference on learning representations},
  volume={2024},
  pages={2632--2652},
  year={2024}
}

@article{jin2021smallholder,
  title={Smallholder cashew plantations in benin},
  author={Jin, Z and Lin, C and Weigl, C and Obarowski, J and Hale, D},
  journal={Radiant MKHub},
  year={2021}
}

@article{helber2019eurosat,
  title={Eurosat: A novel dataset and deep learning benchmark for land use and land cover classification},
  author={Helber, Patrick and Bischke, Benjamin and Dengel, Andreas and Borth, Damian},
  journal={IEEE Journal of Selected Topics in Applied Earth Observations and Remote Sensing},
  volume={12},
  number={7},
  pages={2217--2226},
  year={2019},
  publisher={IEEE}
}

@article{jakubik2023foundation,
  title={Foundation models for generalist geospatial artificial intelligence},
  author={Jakubik, Johannes and Roy, Sujit and Phillips, CE and Fraccaro, Paolo and Godwin, Denys and Zadrozny, Bianca and Szwarcman, Daniela and Gomes, Carlos and Nyirjesy, Gabby and Edwards, Blair and others},
  journal={arXiv preprint arXiv:2310.18660},
  year={2023}
}

@article{irvin2020forestnet,
  title={Forestnet: Classifying drivers of deforestation in indonesia using deep learning on satellite imagery},
  author={Irvin, Jeremy and Sheng, Hao and Ramachandran, Neel and Johnson-Yu, Sonja and Zhou, Sharon and Story, Kyle and Rustowicz, Rose and Elsworth, Cooper and Austin, Kemen and Ng, Andrew Y},
  journal={arXiv preprint arXiv:2011.05479},
  year={2020}
}

@inproceedings{jakubik2025terramind,
  title={Terramind: Large-scale generative multimodality for earth observation},
  author={Jakubik, Johannes and Yang, Felix and Blumenstiel, Benedikt and Scheurer, Erik and Sedona, Rocco and Maurogiovanni, Stefano and Bosmans, Jente and Dionelis, Nikolaos and Marsocci, Valerio and Kopp, Niklas and others},
  booktitle={2025 IEEE/CVF International Conference on Computer Vision (ICCV)},
  pages={7383--7394},
  year={2025},
  organization={IEEE}
}

@article{marsocci2024pangaea,
  title={Pangaea: A global and inclusive benchmark for geospatial foundation models},
  author={Marsocci, Valerio and Jia, Yuru and Bellier, Georges Le and Kerekes, David and Zeng, Liang and Hafner, Sebastian and Gerard, Sebastian and Brune, Eric and Yadav, Ritu and Shibli, Ali and others},
  journal={arXiv preprint arXiv:2412.04204},
  year={2024}
}

@misc{garnot2020lightweighttemporalselfattentionclassifying,
      title={Lightweight Temporal Self-Attention for Classifying Satellite Image Time Series}, 
      author={Vivien Sainte Fare Garnot and Loic Landrieu},
      year={2020},
      eprint={2007.00586},
      archivePrefix={arXiv},
      primaryClass={cs.CV},
      url={https://arxiv.org/abs/2007.00586}, 
}

@inproceedings{xiao2018unified,
  title={Unified perceptual parsing for scene understanding},
  author={Xiao, Tete and Liu, Yingcheng and Zhou, Bolei and Jiang, Yuning and Sun, Jian},
  booktitle={European conference on computer vision},
  pages={432--448},
  year={2018},
  organization={Springer}
}

@article{szwarcman2025prithvi,
  title={Prithvi-eo-2.0: A versatile multi-temporal foundation model for earth observation applications},
  author={Szwarcman, Daniela and Roy, Sujit and Fraccaro, Paolo and G{\'\i}slason, Orsteinn El{\'\i} and Blumenstiel, Benedikt and Ghosal, Rinki and De Oliveira, Pedro Henrique and de Sousa Almeida, Joao Lucas and Sedona, Rocco and Kang, Yanghui and others},
  journal={IEEE Transactions on Geoscience and Remote Sensing},
  year={2025},
  publisher={IEEE}
}

@inproceedings{thomas2019kpconv,
  title={Kpconv: Flexible and deformable convolution for point clouds},
  author={Thomas, Hugues and Qi, Charles R and Deschaud, Jean-Emmanuel and Marcotegui, Beatriz and Goulette, Fran{\c{c}}ois and Guibas, Leonidas J},
  booktitle={Proceedings of the IEEE/CVF international conference on computer vision},
  pages={6411--6420},
  year={2019}
}

@inproceedings{heo2024rotary,
  title={Rotary position embedding for vision transformer},
  author={Heo, Byeongho and Park, Song and Han, Dongyoon and Yun, Sangdoo},
  booktitle={European Conference on Computer Vision},
  pages={289--305},
  year={2024},
  organization={Springer}
}

@inproceedings{Turckheim_2025_BMVC,
author    = {Hugo Riffaud de Turckheim and Diego Marcos and Roberto Interdonato and Sylvain Lobry},
title     = {Atomizer: Generalizing to unseen modalities by breaking images down to a set of scalars},
booktitle = {36th British Machine Vision Conference 2025, {BMVC} 2025, Sheffield, UK, November 24-27, 2025},
publisher = {BMVA},
year      = {2025},
url       = {https://bmva-archive.org.uk/bmvc/2025/assets/papers/Paper_1058/paper.pdf}
}

@inproceedings{perron2026universat,
  title     = {UniverSat: Resolution- and Modality-Agnostic Transformers for Earth Observation},
  author    = {Perron, Yohann and Astruc, Guillaume and Gonthier, Nicolas and Mallet, Clement and Landrieu, Loic},
  booktitle = {Advances in Neural Information Processing Systems},
  year      = {2026},
  note      = {To appear}
}

@inproceedings{sumbul2025smarties,
  title={SMARTIES: Spectrum-aware multi-sensor auto-encoder for remote sensing images},
  author={Sumbul, Gencer and Xu, Chang and Dalsasso, Emanuele and Tuia, Devis},
  booktitle={2025 IEEE/CVF International Conference on Computer Vision (ICCV)},
  pages={5569--5578},
  year={2025},
  organization={IEEE}
}

@inproceedings{prexl2024senpa,
  title={Senpa-mae: Sensor parameter aware masked autoencoder for multi-satellite self-supervised pretraining},
  author={Prexl, Jonathan and Schmitt, Michael},
  booktitle={DAGM German Conference on Pattern Recognition},
  pages={317--331},
  year={2024},
  organization={Springer}
}

@article{tseng2023lightweight,
  title={Lightweight, pre-trained transformers for remote sensing timeseries},
  author={Tseng, Gabriel and Cartuyvels, Ruben and Zvonkov, Ivan and Purohit, Mirali and Rolnick, David and Kerner, Hannah},
  journal={arXiv preprint arXiv:2304.14065},
  year={2023}
}

@article{tseng2025galileo,
  title={Galileo: Learning Global and Local Features in Pretrained Remote Sensing Models},
  author={Tseng, Gabriel and Fuller, Anthony and Reil, Marlena and Herzog, Henry and Beukema, Patrick and Bastani, Favyen and Green, James R and Shelhamer, Evan and Kerner, Hannah and Rolnick, David},
  journal={arXiv preprint arXiv:2502.09356},
  year={2025}
}

@inproceedings{jaegle2021perceiver,
  title={Perceiver: General perception with iterative attention},
  author={Jaegle, Andrew and Gimeno, Felix and Brock, Andy and Vinyals, Oriol and Zisserman, Andrew and Carreira, Joao},
  booktitle={International conference on machine learning},
  pages={4651--4664},
  year={2021},
  organization={PMLR}
}

@inproceedings{reed2023scale,
  title={Scale-mae: A scale-aware masked autoencoder for multiscale geospatial representation learning},
  author={Reed, Colorado J and Gupta, Ritwik and Li, Shufan and Brockman, Sarah and Funk, Christopher and Clipp, Brian and Keutzer, Kurt and Candido, Salvatore and Uyttendaele, Matt and Darrell, Trevor},
  booktitle={Proceedings of the IEEE/CVF International Conference on Computer Vision},
  pages={4088--4099},
  year={2023}
}

@article{astruc2024anysat,
  title={AnySat: An Earth Observation Model for Any Resolutions, Scales, and Modalities},
  author={Astruc, Guillaume and Gonthier, Nicolas and Mallet, Clement and Landrieu, Loic},
  journal={arXiv preprint arXiv:2412.14123},
  year={2024}
}

@article{li2025fleximo,
  title={FlexiMo: A Flexible Remote Sensing Foundation Model},
  author={Li, Xuyang and Li, Chenyu and Ghamisi, Pedram and Hong, Danfeng},
  journal={arXiv preprint arXiv:2503.23844},
  year={2025}
}

@article{xiong2024neural,
  title={Neural plasticity-inspired multimodal foundation model for earth observation},
  author={Xiong, Zhitong and Wang, Yi and Zhang, Fahong and Stewart, Adam J and Hanna, Jo{\"e}lle and Borth, Damian and Papoutsis, Ioannis and Saux, Bertrand Le and Camps-Valls, Gustau and Zhu, Xiao Xiang},
  journal={arXiv preprint arXiv:2403.15356},
  year={2024}
}

@article{jaegle2021perceiverio,
  title={Perceiver io: A general architecture for structured inputs \& outputs},
  author={Jaegle, Andrew and Borgeaud, Sebastian and Alayrac, Jean-Baptiste and Doersch, Carl and Ionescu, Catalin and Ding, David and Koppula, Skanda and Zoran, Daniel and Brock, Andrew and Shelhamer, Evan and others},
  journal={arXiv preprint arXiv:2107.14795},
  year={2021}
}

@inproceedings{waldmann2025panopticon,
  title={Panopticon: Advancing any-sensor foundation models for earth observation},
  author={Waldmann, Leonard and Shah, Ando and Wang, Yi and Lehmann, Nils and Stewart, Adam and Xiong, Zhitong and Zhu, Xiao Xiang and Bauer, Stefan and Chuang, John},
  booktitle={Proceedings of the Computer Vision and Pattern Recognition Conference},
  pages={2204--2214},
  year={2025}
}

@inproceedings{lee2019set,
  title={Set transformer: A framework for attention-based permutation-invariant neural networks},
  author={Lee, Juho and Lee, Yoonho and Kim, Jungtaek and Kosiorek, Adam and Choi, Seungjin and Teh, Yee Whye},
  booktitle={International conference on machine learning},
  pages={3744--3753},
  year={2019},
  organization={PMLR}
}

@InProceedings{Houdre_2026_CVPR,
    author    = {Houdr\'e, Nicolas and Marcos, Diego and de Turckheim, Hugo Riffaud and Ienco, Dino and Wendling, Laurent and Kurtz, Camille and Lobry, Sylvain},
    title     = {RAMEN: Resolution-Adjustable Multimodal Encoder for Earth Observation},
    booktitle = {Proceedings of the IEEE/CVF Conference on Computer Vision and Pattern Recognition (CVPR)},
    month     = {June},
    year      = {2026},
    pages     = {27838-27848}
}

@article{lecun1989cnn,
  title={Backpropagation applied to handwritten zip code recognition},
  author={LeCun, Yann and Boser, Bernhard and Denker, John S and Henderson, Donnie and Howard, Richard E and Hubbard, Wayne and Jackel, Lawrence D},
  journal={Neural Computation},
  year={1989}
}

@article{guo2019review3d,
  title={Deep learning for 3d point clouds: A survey},
  author={Guo, Yulan and Wang, Hanyun and Hu, Qingyong and Liu, Hao and Liu, Li and Bennamoun, Mohammed},
  journal={arXiv preprint arXiv:1912.12033},
  year={2019}
}

\appendix
\newpage
\appendix

\section{Supplementary Material}

\begin{figure}[ht]
  \centering
  \vspace{-0.25cm}
  \includegraphics[width=1.0\textwidth]{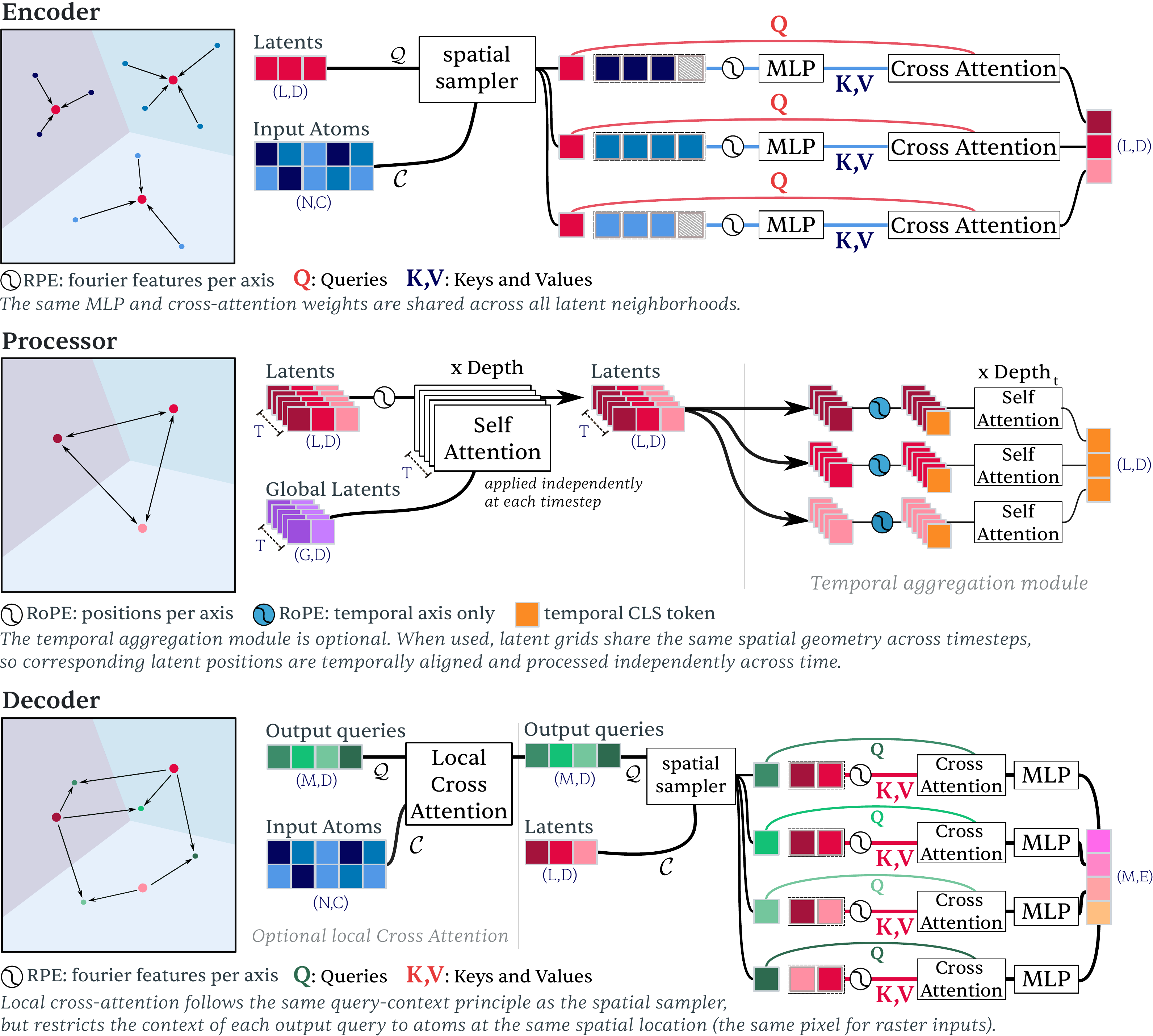}
  \vspace{-0.50cm}
  \caption{
Detailed Atomizer-IO architecture. The encoder aggregates local neighborhoods
of input atoms into spatial latents through the spatial sampler; the processor
applies latent self-attention with optional temporal aggregation; and the
decoder maps local latent neighborhoods to spatial output queries, optionally
conditioning first on raw observations at the target location. Relative
positional information is introduced locally, and MLP and attention weights
are shared across spatial neighborhoods.
}
  \label{fig:whole_arch_appendix}
\end{figure}

\section{Ablation Studies}
\label{app:ablations}

We examine three complementary sources of structure in Atomizer-IO: temporal
organization across acquisitions, spatial relationships between latents, and
direct target-local observations in the decoder. These ablations are evaluated
across tasks with different sensing characteristics to assess how strongly each
source of information contributes to prediction.

\subsection{Temporal Aggregation}
\label{app:temporal_ablation}
\vspace{-0.15cm}
The temporal aggregation module changes how observations across time are
organized in the latent representation. Without it, atoms from all acquisitions are encoded jointly into a single temporally mixed latent set, with acquisition time provided as metadata. With temporal aggregation, each acquisition is encoded independently into its own spatial latent set, and corresponding latents are merged across time by a temporal transformer.

Both conditions process the same total number of atoms; the difference is how the resulting latent capacity is organized. In the joint formulation, each latent receives information from observations spanning both space and time. With temporal aggregation, spatial encoding is performed independently within each acquisition before aligned latent representations are combined temporally.

We evaluate this structural difference on PASTIS by varying the number of input acquisitions while keeping the remaining training setup fixed.

\begin{wrapfigure}[17]{l}{0.60\textwidth}
    \centering
    \vspace{-0.25cm}
    \includegraphics[width=0.47\textwidth]{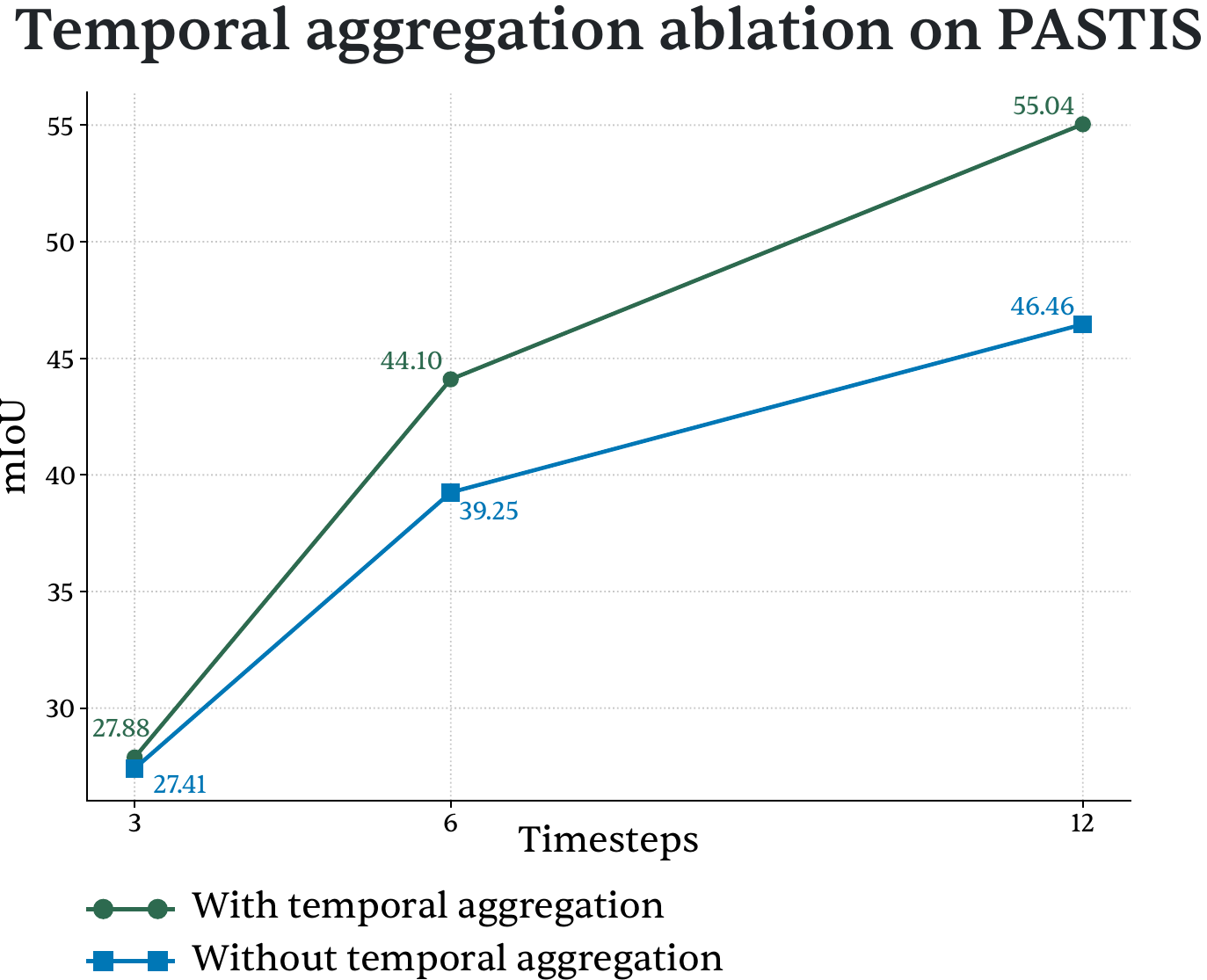}
    \caption{
    Temporal aggregation ablation on PASTIS as the number of input timesteps
    increases.
    }
    \label{fig:temporal_ablation}
\end{wrapfigure}

Explicit temporal aggregation becomes increasingly beneficial as the number of acquisitions grows. With three timesteps, it improves PASTIS by only $0.47$ mIoU, compared with $4.85$ mIoU at six timesteps and $8.58$ mIoU at twelve. Since both conditions process the same observations and the latent capacity scales with the number of input atoms, the growing gap is better explained by how that capacity is structured than by its overall size. Encoding each acquisition spatially before aggregating aligned latents across time provides an increasingly useful inductive bias as the temporal context grows.

\subsection{Spatial RoPE Ablation}
\label{app:rope_ablation}
\vspace{-0.125cm}

We ablate the two-dimensional rotary positional encoding used in the latent
processor while keeping the remaining architecture unchanged. This isolates
the contribution of explicit spatial structure within latent self-attention.

\begin{table}[H]
\vspace{-0.15cm}
\centering
\caption{
Effect of spatial RoPE across tasks. MNIST reports accuracy (\%), while the EO
datasets report mIoU (\%).
}
\label{tab:rope_ablation}
\small
\setlength{\tabcolsep}{6pt}
\renewcommand{\arraystretch}{1.05}
\begin{tabular}{@{}lccccc@{}}
\toprule
& \textbf{MNIST}
& \textbf{Cashew}
& \textbf{PASTIS}
& \textbf{BurnScars}
& \textbf{Sen1Floods11} \\
\midrule
No RoPE
& 29.48
& 62.65
& 40.65
& 86.16
& 92.42 \\

RoPE
& \textbf{99.37}
& \textbf{72.01}
& \textbf{44.10}
& \textbf{88.81}
& \textbf{93.17} \\
\bottomrule
\end{tabular}
\end{table}

\begin{wrapfigure}[17]{r}{0.52\textwidth}
    \vspace{-0.55cm}
    \centering
    \includegraphics[width=0.50\textwidth]{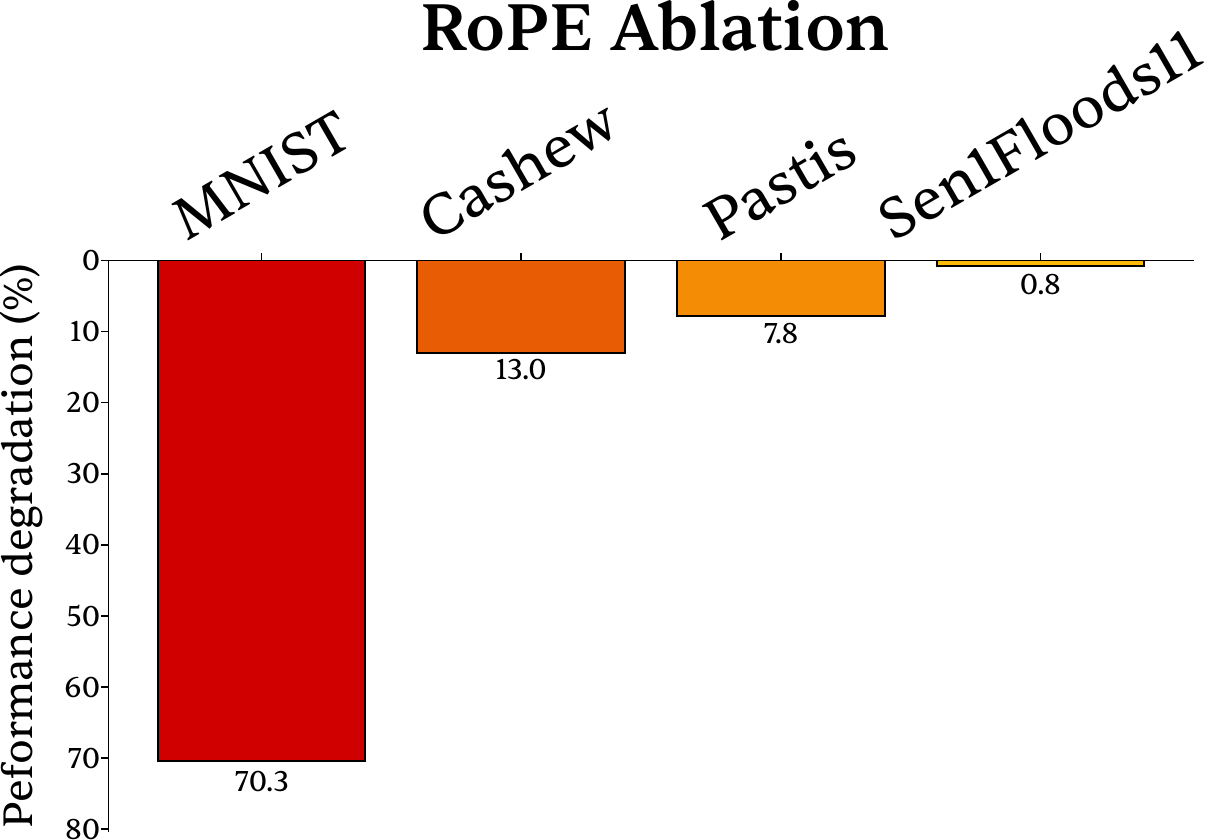}
    \caption{
    Relative degradation when spatial RoPE is removed, computed as
    $100(\mathrm{RoPE}-\mathrm{No~RoPE})/\mathrm{RoPE}$.
    }
    \label{fig:rope_ablation}
\end{wrapfigure}

The contribution of spatial RoPE depends strongly on how informative the local
observations already are. On MNIST, individual pixel values carry little
semantic information in isolation, so the model must rely heavily on spatial
relationships between latents; removing RoPE therefore causes a $70.3\%$
relative performance drop. On Cashew, multispectral measurements provide
useful local cues, but spatial context remains important, leading to a
substantial $13.0\%$ degradation without RoPE. The effect is more moderate on
PASTIS ($7.8\%$) and BurnScars ($3.0\%$), and small on Sen1Floods11 ($0.8\%$),
where local spectral and radar observations already provide strong evidence
for the prediction.

PASTIS provides an additional contrast: spatial RoPE gives a moderate gain,
while the temporal aggregation ablation in
Appendix~\ref{app:temporal_ablation} shows a much larger dependence on temporal
structure. Together, these results suggest that the usefulness of explicit
positional structure depends on which physical relationships are most
informative for the task.

\subsection{Local Observation Conditioning}
\label{app:local_observation_ablation}

For dense prediction, Atomizer-IO can condition each output query directly on
the raw observations at its target location before attending to the latent
representation. We ablate this pathway while keeping the remainder of the
architecture unchanged.

\begin{table}[H]
\centering
\caption{
Effect of conditioning output queries on the raw observations at their target
location. All values are mIoU (\%).
}
\label{tab:local_observation_ablation}
\small
\setlength{\tabcolsep}{7pt}
\renewcommand{\arraystretch}{1.05}
\begin{tabular}{@{}lcccc@{}}
\toprule
& \textbf{Sen1Floods11}
& \textbf{Cashew}
& \textbf{PASTIS}
& \textbf{BurnScars} \\
\midrule
Without local observations
& 89.26
& 70.58
& 43.30
& 88.20 \\

Full model
& \textbf{93.17}
& \textbf{72.01}
& \textbf{44.10}
& \textbf{88.81} \\
\bottomrule
\end{tabular}
\end{table}

\begin{wrapfigure}[18]{l}{0.52\textwidth}
    \centering
    \includegraphics[width=0.50\textwidth]{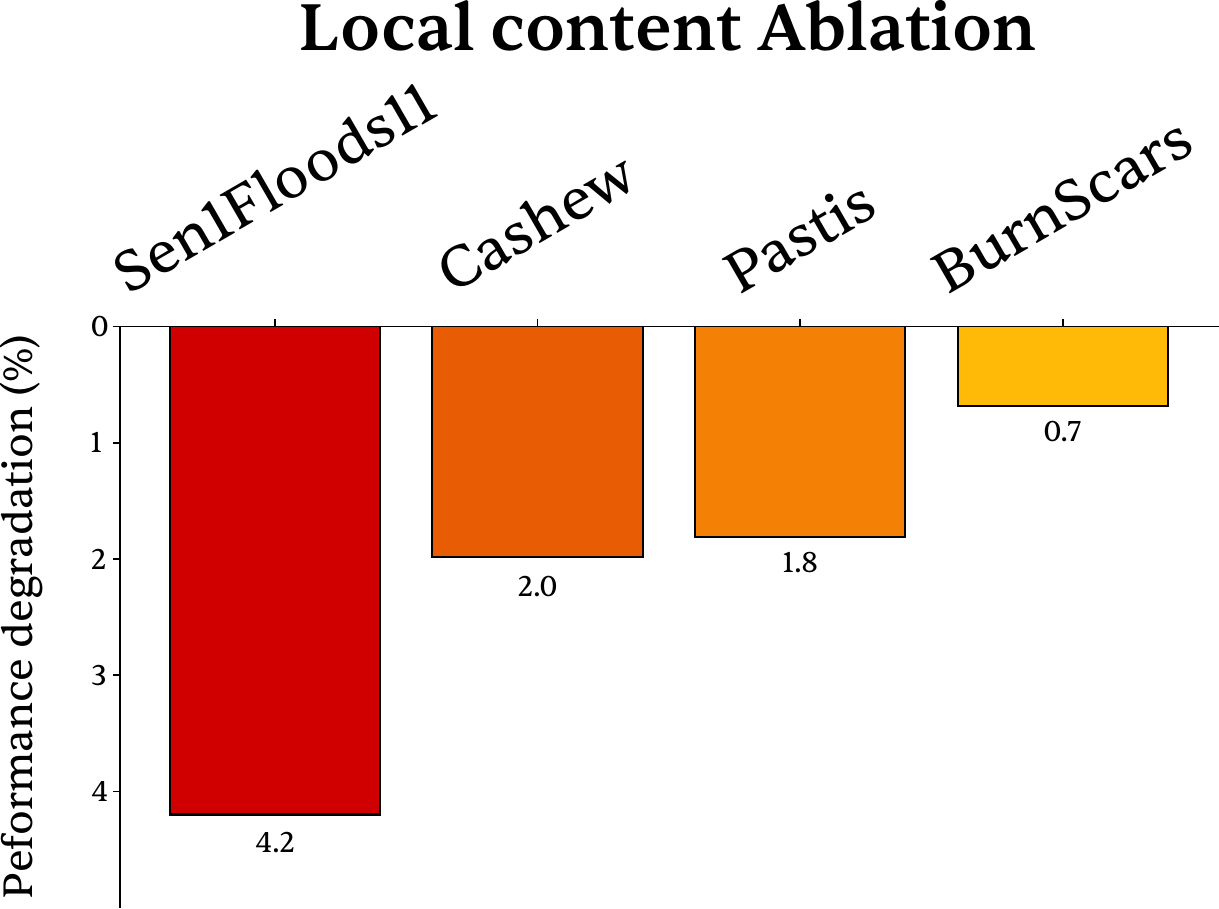}
    \caption{
    Relative degradation when local observation conditioning is removed,
    computed as
    $100(\mathrm{Full}-\mathrm{No~Local})/\mathrm{Full}$.
    }
    \label{fig:skip_ablation}
\end{wrapfigure}

The contribution of direct local observations varies across tasks. Removing
this pathway reduces Sen1Floods11 by $3.91$ mIoU points and Cashew by $1.43$
points. The reductions are smaller on PASTIS ($0.80$ points) and BurnScars
($0.61$ points). Thus, Sen1Floods11 shows the greatest dependence on
target-local evidence among these datasets.

Together with the spatial and temporal ablations, these results show that the
relative importance of target-local evidence, spatial relationships, and
temporal organization is task dependent. PASTIS provides a complementary example: it benefits strongly from explicit temporal aggregation while showing only a modest dependence on direct
target-local observations.

\subsection{Atomic Feature Encoding}
\label{app:token_encoding}

We detail the feature encodings used to build the atomic representation
$\mathbf{f}$ introduced in Section~\ref{sec:token_construction}.
For a raster atom, $(\phi_0,\phi_1,\phi_2,\phi_3)$ are the encoders
$(\phi_v,\phi_g,\phi_\lambda,\phi_t)$ defined below. Unless stated
otherwise, continuous quantities are encoded with Fourier features. For a
scalar $w$, we define
\[
\mathrm{fourier}(w;F,\nu_{\max})
=
\left[~
w,~
\sin(\pi \nu_1 w),\cos(\pi \nu_1 w),~
\ldots,~
\sin(\pi \nu_F w),\cos(\pi \nu_F w)~
\right],
\]
where the $F$ frequencies $\nu_i$ are linearly spaced between $1$ and
$\nu_{\max}$. The encoding has dimension $2F+1$.

\paragraph{Measurement.}
The measured value $v$ is encoded as
\[
\phi_v(v)
=
\mathrm{fourier}(v;F_v,\nu_{\max}^{v}).
\]

\paragraph{Spatial scale.}
For raster observations, spatial scale is represented by the ground sampling distance (GSD) $g$. We first compress it as
\[
\tilde g
=
\mathrm{compress}(g, s_g),
\]
where $s_g$ is a reference GSD, and then encode
\[
\phi_g(g)
=
\mathrm{fourier}(\tilde g;F_g,\nu_{\max}^{g}).
\]

\paragraph{Spectral configuration.}
Following Atomizer~\citep{Turckheim_2025_BMVC}, optical channels are
represented from their physical spectral support rather than a discrete
channel index. For a band with central wavelength $\lambda$ and bandwidth
$\Delta\lambda$,
\[
\phi_\lambda(\lambda,\Delta\lambda)_i
=
\int_{\lambda-\Delta\lambda/2}^{\lambda+\Delta\lambda/2}
\mathcal{N}(\lambda';\mu_i,\sigma_i)\,d\lambda',
\qquad i=1,\ldots,K,
\]
where $\mu_i$ and $\sigma_i$ are the center and width of the $i$-th Gaussian
basis function. The resulting vector depends on both central wavelength and
bandwidth. Channels without meaningful spectral support, such as SAR
polarizations or elevation, use learned embeddings of the same dimensionality.

\paragraph{Acquisition time.}
Acquisition time is represented by day of year $t$ using periodic Fourier
features,
\[
\phi_t(t)
=
\left[
\sin\!\left(\frac{2\pi k t}{365}\right),
\cos\!\left(\frac{2\pi k t}{365}\right)
\right]_{k=1}^{K_t}.
\]
This gives a $2K_t$-dimensional encoding with annual periodicity. When
acquisition time is unavailable, a zero vector is used.

\subsection{Relative Position Encoding}
\label{app:rpe_encoding}
Horizontal spatial coordinates are kept separate from $\mathbf{f}$ and enter through the offset $\boldsymbol{\delta}$ between a context element and its query. Each component $\delta$ is encoded as
\[
\phi_\delta(\delta) = \mathrm{fourier}\big(\mathrm{compress}(\delta, s_0);, F_\delta,, \nu_{\max}^{\delta}\big),
\]
and the component encodings are concatenated. 
The reference scale $s_0$ is the cross-attention spatial scale of Table~\ref{tab:atomizer_hyperparameters}, and $F_\delta$, $\nu_{\max}^{\delta}$ are in 
Table~\ref{tab:fourier_parameters}.

\subsection{Properties of the Compression Function}
\label{app:compression}

For physical quantities spanning a large range, we use
\[
\mathrm{compress}(w,s)=\frac{w}{s+|w|},
\]
where $s>0$ is a reference scale. The function maps signed inputs to
$(-1,1)$ and non-negative inputs to $[0,1)$. Near zero,
$\mathrm{compress}(w,s)\approx w/s$, and its magnitude saturates toward one
for large $|w|$. In particular, $|\mathrm{compress}(w,s)|=0.5$ when $|w|=s$.
Thus, $s$ sets the characteristic physical scale before Fourier encoding. Separate
reference scales are used for GSD and relative spatial positions.
\subsection{Scalability of the Spatial Sampler}
\label{app:spatial_sampler_complexity}

We analyze the encoder instantiation of the spatial sampler, where each input
atom is assigned to its nearest spatial latent. This Voronoi assignment replaces
global input-to-latent cross-attention with local cross-attention, reducing the
number of query--context interactions. During training, the context available
to each latent can additionally be subsampled to a user-defined budget $m$,
providing direct control over cross-attention cost.

Table~\ref{tab:complexity} compares this local formulation with global
cross-attention for the Sen1Floods11 configuration used in our experiments:
15 channels at $512\times512$ resolution
($N\approx3.9\times10^6$ atoms) and $L=2000$ spatial latents.

Without subsampling, every atom participates in exactly one encoder
cross-attention neighborhood, so the Voronoi assignment reduces the number of
input-to-latent interactions from $LN$ to $N$. With a per-latent budget $m$,
the number of interactions becomes
$\sum_{\ell=1}^{L}\min(m,|V_\ell|)$ and is therefore bounded by
$\min(N,Lm)$.

For comparison, we report the \emph{equivalent global drop rate}, $1-m/N$:
the fraction of atoms that would need to be removed from the context of each
global latent for global cross-attention to use at most $m$ context elements
per latent. For $m=500$, this corresponds to $99.987\%$: each latent in a
global formulation would retain only approximately 500 atoms out of
$3.9$ million. The spatial sampler instead distributes interactions spatially:
every atom remains assigned to a local cell, although dense cells may be
subsampled at a given training step.

\begin{table}[ht]
\centering
\caption{
Cross-attention complexity for global and local spatial sampling on
Sen1Floods11 ($N\approx3.9\times10^6$ atoms, $L=2000$ spatial latents).
The equivalent global drop rate is $1-m/N$, the fraction of atoms that global
cross-attention would need to discard to match a budget of $m$ context elements
per latent.
}
\label{tab:complexity}
\begin{tabular}{lccc}
\toprule
& \textbf{Interaction bound}
& \textbf{Equiv.\ global drop}
& \textbf{Context/query} \\
\midrule
Global
& $LN \approx 7.86\times10^9$
& 0\%
& $N$ \\
Local ($m{=}500$)
& $\leq Lm = 1.0\times10^6$
& 99.987\%
& $\leq 500$ \\
Local ($m{=}1000$)
& $\leq Lm = 2.0\times10^6$
& 99.975\%
& $\leq 1000$ \\
Local ($m{=}2000$)
& $\leq \min(N,Lm) \approx 3.93\times10^6$
& 99.949\%
& $\leq 2000$ \\
\bottomrule
\end{tabular}
\end{table}

The budget $m$ is not an architectural limit on local context. The complete
context available to latent $\ell$ is its Voronoi cell $V_\ell$. If
$|V_\ell|\leq m$, the full cell is used; otherwise, $m$ controls how much of
that context is processed at a given training step. For $L=2000$, spatial
latent self-attention contributes $L^2=4\times10^6$ pairwise interactions,
compared with $7.86\times10^9$ for global input-to-latent cross-attention.

When a cell contains more atoms than the selected budget $m$, atoms are sampled
uniformly without replacement at each training step. This controls training
cost without permanently restricting a latent to a fixed subset of its local
context. Implementation details are given in
Appendix~\ref{app:spatial_sampler_implementation}, and sensitivity to $m$ is
reported in Appendix~\ref{fig:Ablat_GLFOPS}.

When the input and latent geometries are fixed, the Voronoi assignment depends
only on their spatial positions and can be precomputed and cached. The only
per-step operation is then the optional subsampling of each local cell.

\vspace{-0.25cm}
\subsection{Implementation Details of the Spatial Sampler}
\label{app:spatial_sampler_implementation}

We detail three implementation steps for the encoder: Voronoi assignment,
per-step subsampling, and masking of variable-size local contexts.

\vspace{-0.2cm}
\paragraph{Voronoi partitioning.}
For each atom $i$ at position $\mathbf{p}_i\in\mathbb{R}^2$, we store its
nearest spatial latent,
\[
\ell(i)
=
\mathrm{kNN}_{k=1}
\left(
\mathbf{p}_i,
\{\boldsymbol{\mu}_{\ell'}\}_{\ell'}
\right),
\]
where $\boldsymbol{\mu}_{\ell'}\in\mathbb{R}^2$ denotes the position of latent
$\ell'$. We then invert this assignment to obtain the complete cell associated
with latent $\ell$,
\[
V_\ell
=
\{i:\ell(i)=\ell\}.
\]
The complete local context available to latent $\ell$ is therefore determined
by $V_\ell$. Any subsequent subsampling is performed from this full cell.
\vspace{-0.2cm}
\paragraph{Variable latent layouts.}
Because no learned parameter depends on the number or placement of spatial
latents, both can be varied on the fly during training. We precompute the
Voronoi assignments for several latent layouts, corresponding to different
values of $\kappa$ (Appendix~\ref{app:training_details}), and randomly assign
one layout to each batch. Since assignments are computed in advance, varying
the layout adds no cost to the training step. At test time, we evaluate a
single configuration $(\kappa,m)$, selected on the validation set.
\vspace{-0.2cm}
\paragraph{Per-step random subsampling.}
During training, the local context can be limited to a user-defined budget
$m$. At step $t$, we sample uniformly without replacement,
\[
S_\ell^{(t)}
\subseteq
V_\ell,
\qquad
|S_\ell^{(t)}|
=
\min(m,|V_\ell|).
\]
When $m\geq|V_\ell|$, the complete cell is used. Otherwise, $m$ directly
controls the number of local query--context interactions processed at that
step. Repeated sampling exposes the model to different subsets of dense cells
rather than permanently restricting each latent to a fixed subset of atoms.

\vspace{-0.2cm}
\paragraph{Masking.}
Because local contexts can contain different numbers of atoms, they are
batched to a common capacity and invalid slots are suppressed with an
attention bias,
\[
\beta_{\ell i}
=
\begin{cases}
0, & \text{if position } i \text{ is valid},\\
-\infty, & \text{otherwise}.
\end{cases}
\]
After the attention softmax, invalid positions receive zero weight. When a
cell is empty, a dummy token is inserted so that the softmax remains well
defined. The cardinality of $\mathcal{V}_\ell$ determines the available local
context, while $m$ controls how much of that context is processed when
subsampling is used.

\vspace{-0.2cm}
\section{GFLOPs}
\vspace{-0.25cm}
\subsection{Adaptive Decoding Strategies}
\label{app:decode_strategies}
\vspace{-0.25cm}
As in the encoder, Atomizer-IO uses cross-attention in the decoder so that
predictions can be produced at explicitly specified spatial locations. For
dense segmentation, this can mean querying every pixel, while other output
layouts can be queried without changing the architecture.

\textit{Dense} decoding processes every output query independently. For each of the
$O$ output queries, the decoder gathers a local neighborhood of $k$ latents,
computes their relative positional encoding, and applies cross-attention.
The resulting cost therefore scales linearly with the number of decoded
locations.

We exploit the spatial redundancy of dense predictions with two adaptive
decoding strategies \textit{(Zone-probe, Quadtree)} that reduce the number of locations receiving a full
decode. Both operate on the same trained checkpoint as (\emph{Dense})
decoding; they differ only in where output queries are evaluated and where
predictions are propagated spatially.

Both strategies use agreement between a small number of probes as a proxy for
local label homogeneity. In \emph{Zone-probe}, the regions are the Voronoi
cells associated with the spatial latents. A small number of locations is
sampled within each cell; if all probes predict the same class, that class is
assigned to the remaining locations, whereas disagreeing cells are fully
decoded. In \emph{Quadtree} decoding, the same principle is applied
hierarchically. Disagreeing regions are recursively subdivided and probed at
finer spatial scales, while homogeneous regions terminate early and broadcast
their predicted class. A full per-pixel decode is recovered at the finest
scale $s_{\min}=1$.

Probe agreement is a heuristic rather than a guarantee: small or thin objects
may be missed if no probe falls on them. The resulting approximation is
evaluated empirically on Sen1Floods11 and BurnScars in
Figure~\ref{fig:Ablat_GLFOPS}.
\vspace{-0.2cm}
\begin{table}[ht]
\centering
\caption{
Decode cost of the three inference strategies. $M$ is the number of output
queries, $L$ the number of spatial latents, $k_p$ the number of probes per
region, and $s_{\text{start}}$ the initial quadtree cell size. Both adaptive
strategies reduce to dense decoding in the worst case when no region can be
treated as homogeneous.
}
\label{tab:decode_complexity}
\begin{tabular}{lccc}
\toprule
& \textbf{Full decode calls}
& \textbf{Best case}
& \textbf{Worst case} \\
\midrule
Dense
& $M$
& $M$
& $M$ \\
Zone-probe
& $k_p L + \sum_{\ell \in \mathrm{hard}} |W_\ell|$
& $\mathcal{O}(k_p L)$
& $M$ \\
Quadtree
& $\lesssim k_p \times (\text{levels} \times \text{active cells})$
& $\mathcal{O}\!\left(\dfrac{k_p HW}{s_{\text{start}}^2}\right)$
& $M$ \\
\bottomrule
\end{tabular}
\end{table}
\vspace{-0.25cm}
Both adaptive strategies recover the dense prediction wherever a full decode
is issued. Their savings therefore depend on prediction homogeneity at the
scale of a latent's Voronoi cell for zone-probe and an image-plane cell for
quadtree.

\vspace{-0.25cm}
\subsection{Compute--Performance Sensitivity}
\begin{figure}[H]
  \centering
  \vspace{-0.25cm}
  \includegraphics[width=1.0\textwidth]{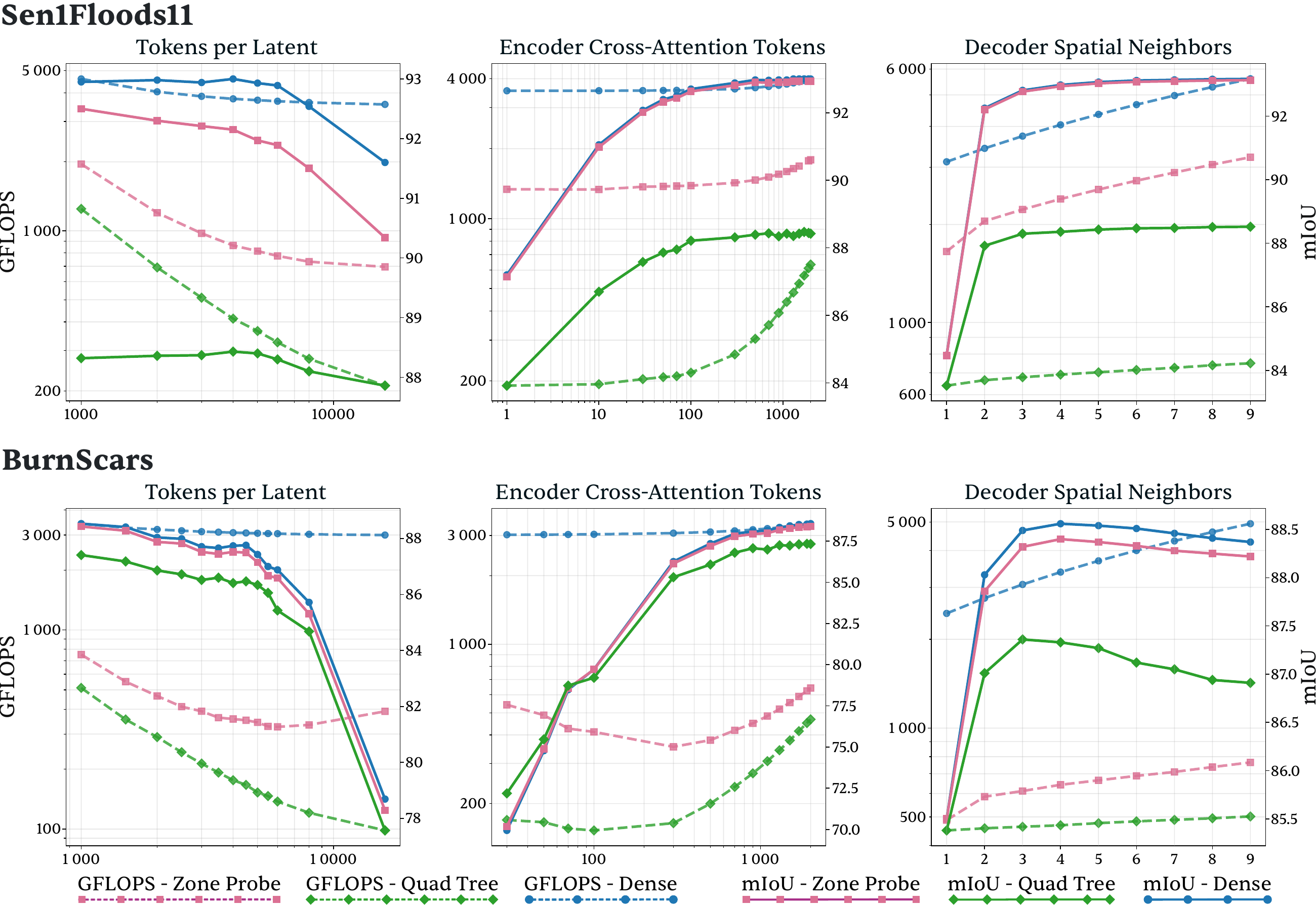}
  \vspace{-0.50cm}
\caption{
Sensitivity of segmentation performance and inference cost to three
architectural parameters: atoms per latent, encoder cross-attention budget,
and decoder neighborhood size. Results are shown on Sen1Floods11 and
BurnScars for Dense, Zone-probe, and Quadtree decoding.
}
  \label{fig:Ablat_GLFOPS}
\end{figure}

\vspace{-0.25cm}
Figure~\ref{fig:Ablat_GLFOPS} examines how the main architectural and
decoding parameters affect both computational cost and segmentation
performance on Sen1Floods11 and BurnScars. We vary the number of atoms per
latent, the encoder cross-attention budget, and the decoder spatial
neighborhood size for dense, zone-probe, and quadtree decoding. For all strategies we use the same trained checkpoint.

Increasing the number of atoms per latent reduces the number of spatial
latents and eventually degrades mIoU across all three decoding strategies.
Because the same trend appears independently of the decoding policy, the loss
reflects reduced representational capacity in the latent representation rather
than a decoder-specific approximation. Increasing the encoder cross-attention
budget improves performance until saturation, while increasing the decoder
neighborhood size increases computation once sufficient local
context is available.

The two adaptive decoders also have different computational floors.
Zone-probe is coupled to the number of spatial latents $L$: reducing its cost
through a lower latent density eventually also reduces encoder capacity.
Quadtree instead controls how densely the existing latent representation is
read out through $s_{\text{start}}$, leaving the representation itself
unchanged. It can therefore reach more aggressive low-cost operating points
before representational quality is affected. The two strategies are thus
complementary: zone-probe covers the regime closest to dense decoding, while
quadtree extends toward stronger compute reduction.
\vspace{-0.15cm}

\section{Echo Encoding for LiDAR Tokens}
\label{app:echo-encoding}

Airborne LiDAR sensors may record multiple returns from a single emitted
pulse as it interacts with structures along its path. Each return stores two
integer attributes: the \emph{return number} $r$, indicating its order within
the pulse, and the \emph{number of returns} $R$, indicating the total number
of returns associated with that pulse, with $1\leq r\leq R$. Together,
$(r,R)$ describes the position of an observation within the return sequence.
This provides useful information for semantic interpretation: surfaces such
as buildings or ground often produce isolated returns, while vegetation
canopies can produce several returns at different depths along the beam path.

\subsection{Encoding}

We encode each return using two normalized quantities,
\begin{align}
a &= \frac{r-1}{R},
\qquad &&\text{(preceding returns)}
\label{eq:echo-above} \\
b &= \frac{R-r}{R},
\qquad &&\text{(subsequent returns)}
\label{eq:echo-below}.
\end{align}
Both lie in $[0,1)$. The first measures the normalized number of returns
preceding the current observation, while the second measures those following
it.

\begin{figure}[H]
  \centering
  \vspace{-0.25cm}
  \includegraphics[width=1.0\textwidth]{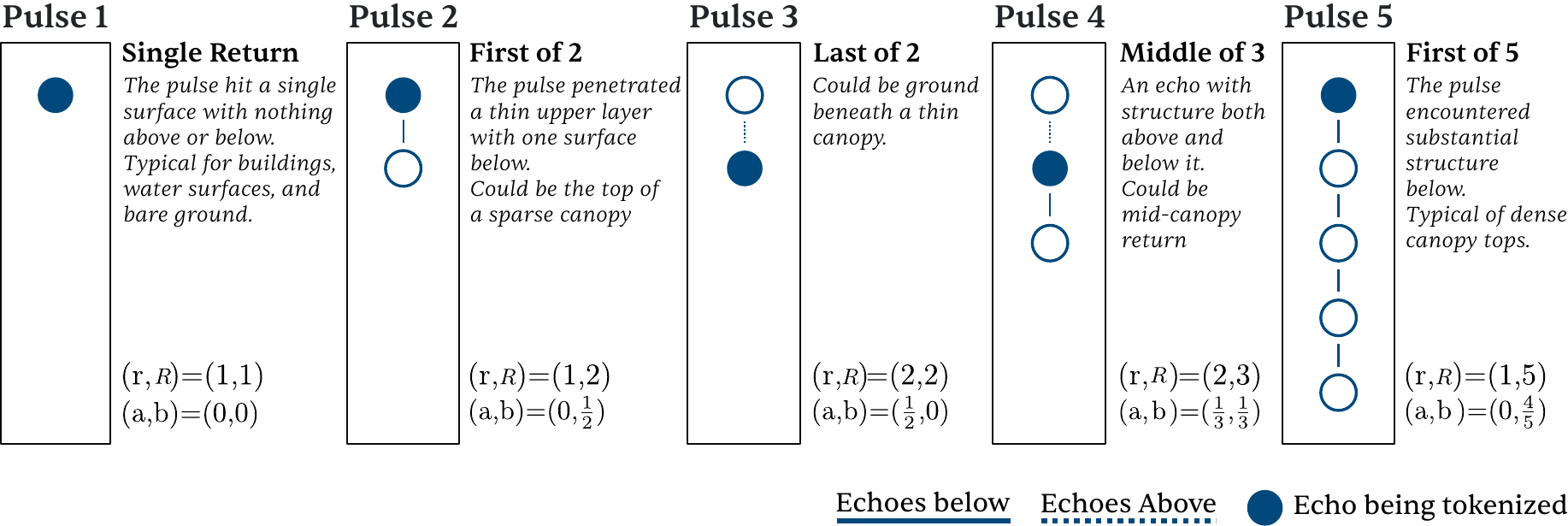}
  \vspace{-0.50cm}
  \caption{
  LiDAR return encoding. The pair $(a,b)$ describes the relative position of
  each observation within its pulse's return sequence. Normalization by the
  per-pulse return count $R$ makes the representation independent of the
  sensor-specific maximum number of returns while preserving the original
  $(r,R)$ attributes.
  }
  \label{fig:App_echoex}
\end{figure}

Unlike normalization by a fixed sensor-specific maximum return count, this
encoding depends only on the returns observed for the current pulse. The same
representation can therefore be used for sensors supporting different maximum
numbers of returns without changing the architecture.

The transformation is also invertible on its valid domain. Since
\[
a+b=\frac{R-1}{R},
\]

the original attributes can be recovered as
\[
R=\frac{1}{1-(a+b)},
\qquad
r=aR+1.
\]

Thus, the transformation preserves all information contained in the original $(r,R)$ pair.

\subsection{Training Details}
\label{app:training_details}

All models are trained from scratch under matched, task-specific training
protocols. Hyperparameters such as batch size and training duration are selected
at the dataset level and shared across all architectures evaluated on that
dataset. Architecture-specific settings required by individual methods are
reported below.

\paragraph{Model capacity.}
Model sizes are selected so that the compared architectures operate in a
similar parameter regime within each task family. We use the ViT-Small
configuration for all Vision Transformer baselines.
Table~\ref{tab:model_parameters} reports the number of trainable parameters
used in the 2D experiments.

\begin{table}[ht]
\centering
\caption{
Number of trainable parameters for the architectures used in the 2D
experiments. Values are given in millions of parameters. ViT refers to
ViT-Small.
}
\label{tab:model_parameters}
\small
\setlength{\tabcolsep}{5.5pt}
\renewcommand{\arraystretch}{1.05}
\begin{tabular}{@{}lccc@{}}
\toprule
\textbf{Architecture}
& \textbf{Classification}
& \textbf{Segmentation}
& \textbf{Multitemporal seg.} \\
\midrule
ResNet50       & 23.6 & 37.3 & 37.3 \\
ViT-Small      & 21.7 & 30.3 & 31.5 \\
Perceiver IO   & 19.7 & 34.5 & 34.5 \\
RAMEN          & 23.1 & 33.7 & 33.7 \\
UniverSat      & 36.1  & 36.1 & 36.1 \\
Atomizer-IO    & 19.2 & 33.5 & 34.8 \\
\bottomrule
\end{tabular}
\end{table}

For Atomizer-IO, the dense single-temporal model contains 33.5M parameters.
Adding the temporal aggregation module increases this to 34.8M, while the
model without temporal aggregation remains at 33.5M. The number of spatial
latents scales with the number of input atoms, but this changes the size of the
intermediate representation rather than the number of learned parameters,
since the same weights are shared across latents.

For context, EO-specific models evaluated in
PANGAEA~\citep{marsocci2024pangaea} typically use approximately
$30$--$47$M trainable parameters. Our dense-prediction models contain
approximately $30$--$37$M parameters and therefore operate in a comparable
capacity regime.

\paragraph{Decoders, temporal aggregation, and classification heads.}
For dense prediction, we follow the PANGAEA
protocol~\citep{marsocci2024pangaea} and equip ResNet50 and ViT with a
UPerNet decoder~\citep{xiao2018unified}. RAMEN natively uses a UPerNet
decoder, while Perceiver-IO and UniverSat use their native decoders. For
multitemporal inputs, we also follow PANGAEA: ResNet50 applies a convolution
before temporal aggregation, and ViT aggregates timesteps with an
L-TAE~\citep{garnot2020lightweighttemporalselfattentionclassifying}. RAMEN,
UniverSat, Perceiver-IO, and Atomizer-IO natively handle the temporal
dimension. For classification, Atomizer-IO and UniverSat use mean pooling
over their latent representations, ViT uses a CLS token, and ResNet50 and
RAMEN use their native classification heads.

\paragraph{Baseline-specific hyperparameters.}
RAMEN and UniverSat require a small number of architecture-specific spatial
hyperparameters. We contacted the authors of both methods to inform the choice
of these settings. For RAMEN, we set the encoding resolution to four times the
native ground sampling distance, except for ForestNet, where we use $40$\,m.
UniverSat uses the same encoding resolutions and a one-pixel subpatch size for
all datasets.

For UniverSat, we follow the authors' recommendation to use the smallest
decoder stride that is computationally tractable. We use a stride of one pixel
for Cashew and PASTIS and a stride of four pixels for the remaining
dense-prediction datasets; for BioMassters in particular, the substantially
larger training set makes the one-pixel configuration computationally
impractical. Classification tasks do not use a decoder. The resulting
configurations are summarized in
Table~\ref{tab:architecture_hyperparameters}.

For Perceiver-IO, we use $512$ latents, one cross-attention layer, and six
self-attention layers. Its metadata encodings use the same dimensions as
those of Atomizer-IO.

\begin{table}[ht]
\centering
\caption{
Architecture-specific spatial hyperparameters used for RAMEN and UniverSat.
Encoding resolutions are expressed in physical units.
}
\label{tab:architecture_hyperparameters}
\small
\setlength{\tabcolsep}{5.5pt}
\renewcommand{\arraystretch}{1.05}
\begin{tabular}{@{}lcccc@{}}
\toprule
&
\textbf{RAMEN}
&
\multicolumn{3}{c}{\textbf{UniverSat}} \\
\cmidrule(lr){2-2}
\cmidrule(lr){3-5}
\textbf{Dataset}
& \textbf{Enc. res. (m)}
& \textbf{Enc. res. (m)}
& \textbf{Subpatch (px)}
& \textbf{Dec. stride (px)} \\
\midrule
ForestNet
& 40
& 40
& 1
& -- \\

BurnScars
& 120
& 120
& 1
& 4 \\

EuroSAT
& 40
& 40
& 1
& -- \\

Cashew
& 40
& 40
& 1
& 1 \\

Sen1Floods11
& 40
& 40
& 1
& 4 \\

xView2
& 2
& 2
& 1
& 4 \\

BioMassters
& 40
& 40
& 1
& 4 \\

PASTIS
& 40
& 40
& 1
& 1 \\
\bottomrule
\end{tabular}
\end{table}

\paragraph{Training protocol.}
All models are optimized with AdamW using a learning rate of $10^{-4}$,
$\beta_1=0.9$, $\beta_2=0.999$, and $\epsilon=10^{-8}$. We use a cosine
annealing learning-rate schedule with a linear warm-up over the first $5\%$ of
training steps. Models are trained for 100 epochs by default. Sen1Floods11 and
BioMassters are trained for 150 epochs, as preliminary experiments showed that
several architectures had not fully converged after 100 epochs. The 3D
experiments on DALES and FRACTAL are trained for 100 epochs. Within each
dataset, the same batch size, learning-rate schedule, and epoch budget are used
for all architectures.

\begin{wraptable}[18]{l}{0.34\textwidth}
\centering

\caption{
Batch size used for each dataset. The same batch size is used across
architectures within a dataset.
}
\label{tab:batch_sizes}
\vspace{-0.15cm}
\small
\setlength{\tabcolsep}{6pt}
\renewcommand{\arraystretch}{1.05}
\begin{tabular}{@{}lc@{}}
\toprule
\textbf{Dataset} & \textbf{Batch size} \\
\midrule
ForestNet      & 8  \\
BurnScars      & 8  \\
EuroSAT        & 8  \\
Cashew         & 8  \\
Sen1Floods11   & 4  \\
xView2         & 8  \\
BioMassters    & 16 \\
PASTIS         & 4  \\
DALES          & 32 \\
FRACTAL        & 20 \\
\bottomrule
\end{tabular}
\vspace{-0.2cm}
\end{wraptable}

\paragraph{Atomizer-IO hyperparameters.}
For the 2D benchmarks, Atomizer-IO uses a latent dimension of $512$, $128$
global latents, and eight attention heads in both cross-attention and
self-attention. Each decoder query attends to its nine nearest spatial
latents. Relative positions used in local cross-attention are normalized with
a fixed spatial scale of $100$\,m. For latent self-attention, the learned
spatial scale is initialized to half the physical extent of the input,
\[
s_p = \frac{g\,n}{2},
\]
where $g$ is the ground sampling distance in meters per pixel and $n$ is the
image size along one spatial dimension. The scale is then optimized jointly
with the rest of the model.

Given $N$ input atoms and a target number of atoms per spatial latent $k$, we
use $L=\lceil N/k\rceil$ spatial latents. Their anchors are placed on a
hexagonal lattice covering the horizontal spatial extent of the input. The
density of the spatial latent representation and the local encoder
cross-attention budget can be randomized jointly during training. For each
batch, we sample one configuration $(k,m)$, where $m$ is the maximum number of
atoms sampled from each latent neighborhood for cross-attention. For PASTIS,
the configurations are
\[
(k,m)\in\{(100,100),(250,250),(350,350)\};
\]
for the other 2D benchmarks, they are
\[
(k,m)\in\{(1000,1000),(1500,1500),(2000,2000)\}.
\]
Varying $k$ changes the number and density of spatial latents without changing
the learned parameters, while varying $m$ changes the local encoder attention
budget.

For dense prediction, we follow the query-subsampling strategy of
Perceiver IO and supervise at most $100{,}000$ spatial output queries per
training sample. When an output contains more than $100{,}000$ locations, the
supervised queries are sampled uniformly at random; otherwise, all locations
are supervised. A $512\times512$ prediction map therefore uses $100{,}000$
sampled queries during training, whereas a $128\times128$ output is supervised
densely.

\begin{table}[ht]
\centering
\caption{
Main Atomizer-IO architectural hyperparameters used for the 2D and 3D
experiments. Here, $k$ denotes the target number of atoms per spatial latent
and $m$ the maximum number of atoms sampled per latent for local
cross-attention.
}
\label{tab:atomizer_hyperparameters}
\small
\setlength{\tabcolsep}{5.5pt}
\renewcommand{\arraystretch}{1.05}
\begin{tabular}{@{}lcc@{}}
\toprule
\textbf{Hyperparameter}
& \textbf{2D}
& \textbf{3D} \\
\midrule

Latent dimension
& 512
& 768 \\

Global latents
& 128
& 128 \\

Attention heads
& 8
& 8 \\

Decoder neighbors
& 9
& 9 \\

Latent layout
& Hexagonal
& Hexagonal \\

$(k,m)$
& \begin{tabular}[c]{@{}l@{}}
Other 2D: $\{(1000,1000),(1500,1500),(2000,2000)\}$ \\
PASTIS: $\{(100,100),(250,250),(350,350)\}$
\end{tabular}
& $(1000,1000)$ \\

Cross-attention spatial scale
& 100\,m
& 100\,m \\

Maximum supervised queries
& $100{,}000$
& $100{,}000$ \\
\bottomrule
\end{tabular}
\end{table}

For the 3D experiments on DALES and FRACTAL, we increase the latent dimension
to $768$ and use six self-attention layers. The model retains $128$ global
latents, eight attention heads, and nine decoder neighbors. Both 3D datasets
use the fixed configuration $(k,m)=(1000,1000)$, with spatial latent anchors
placed on the same hexagonal lattice over the horizontal extent of each input.

\paragraph{Fourier feature encoding.}
The continuous metadata used by Atomizer-IO are encoded with Fourier features
using the parameters reported in Table~\ref{tab:fourier_parameters}. The same
encoding and hyperparameters are used for Perceiver IO, so both models receive
the same continuous metadata representation.

\begin{table}[ht]
\centering
\caption{
Fourier feature parameters used by Atomizer-IO and Perceiver IO.
}
\label{tab:fourier_parameters}
\small
\setlength{\tabcolsep}{7pt}
\renewcommand{\arraystretch}{1.05}
\begin{tabular}{@{}lcc@{}}
\toprule
\textbf{Feature}
& \textbf{Frequency bands}
& \textbf{Maximum frequency / period} \\
\midrule
Position
& 16
& 16 \\

Resolution
& 2
& 2 \\

Acquisition time
& 12
& 365 days \\

Measurement value
& 8
& 8 \\
\bottomrule
\end{tabular}
\end{table}

\subsection{MNIST Sparsification Protocol}
\label{app:mnist_protocol}

\begin{wrapfigure}[16]{r}{0.42\textwidth}
    \centering
    \vspace{-0.35cm}
    \includegraphics[width=0.40\textwidth]{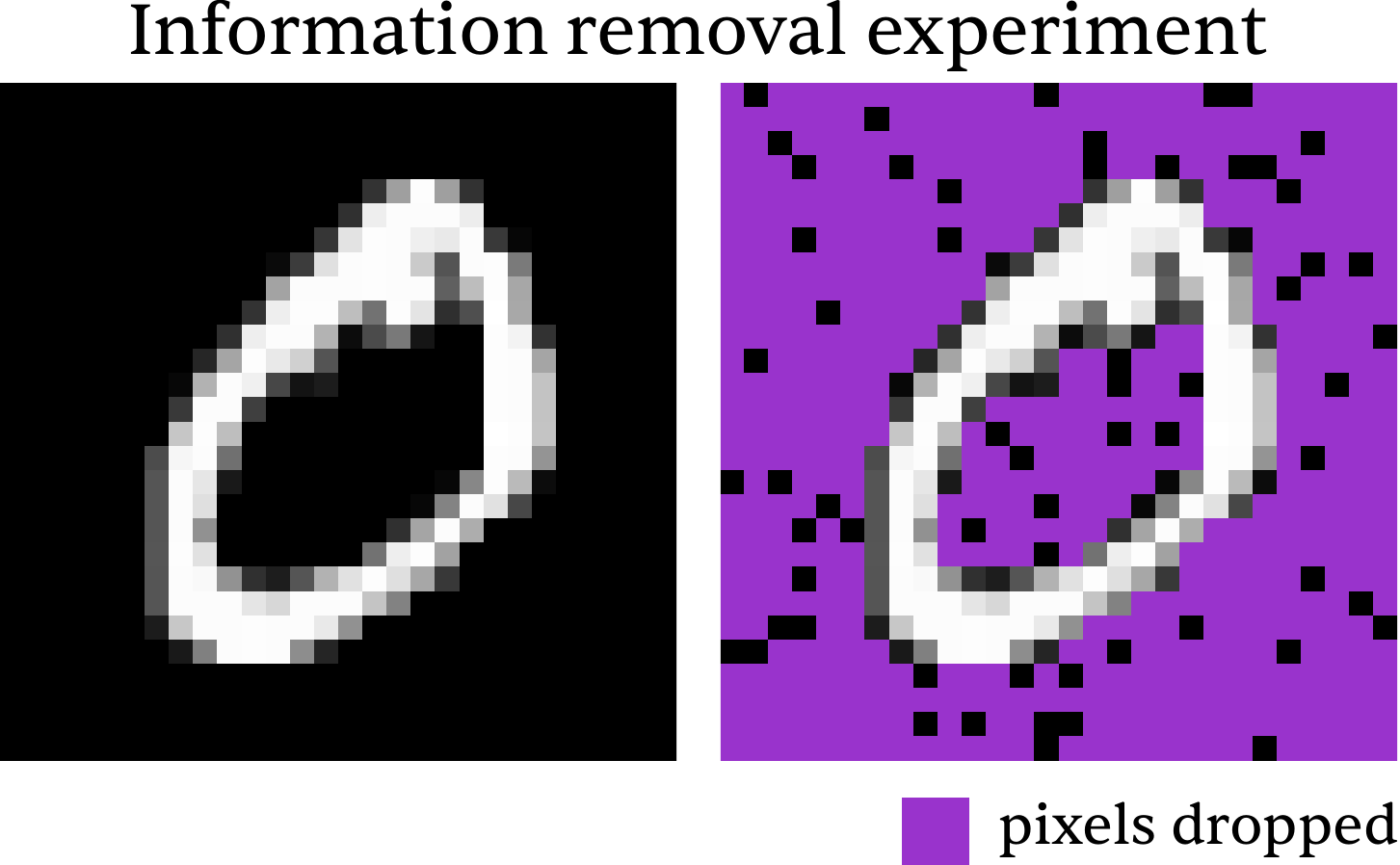}
    \caption{
    MNIST sparsification protocol. Models are trained on complete images.
    At evaluation, pixels with value $v\leq0.5$ are progressively removed,
    while pixels with value $v>0.5$ are always retained.
    }
    \label{fig:mnist_sparsification}
    \vspace{-0.25cm}
\end{wrapfigure}

MNIST is used as a controlled test of spatial reasoning under changes in input
geometry. To control for model capacity on this relatively simple dataset, we
compare a $7.40$M-parameter Atomizer-IO with a parameter-matched
$7.44$M-parameter ViT. The ViT operates on $1\times1$ patches, giving both
models the same pixel-level spatial granularity.

Atomizer-IO uses a latent dimension of $256$, four global latents, four
cross-attention heads, and eight latent self-attention heads. The encoder
contains one cross-attention layer, followed by a processor with four
self-attention layers. Each retained input pixel is associated with one spatial
latent centered at the same location. Spatial RoPE is used only in latent
self-attention; encoder cross-attention receives no relative positional
encoding. Pixel coordinates are expressed in units of one per pixel, and the
learned RoPE scale is initialized to $14$ coordinate units, corresponding to
half the width of the $28\times28$ image.

Both models are trained for 50 epochs with batch size $64$ using AdamW with a
learning rate of $10^{-4}$ and weight decay $10^{-3}$. Training uses complete
MNIST images with no input sparsification.

At evaluation, we progressively remove only pixels with value $v\leq0.5$,
while pixels with $v>0.5$ are always retained
(Figure~\ref{fig:mnist_sparsification}). The same pixels are removed for both
architectures, and removed pixels are absent from the input rather than
replaced by zero-valued tokens. For the ViT, retained tokens keep the learned
absolute positional embeddings associated with their original pixel locations,
so removing pixels does not re-index the remaining observations. Neither model
is retrained or fine-tuned for the resulting sparse input geometries.

\begin{table}[ht]
\centering
\caption{
Architecture and training hyperparameters for the capacity-matched
Atomizer-IO and ViT models used in the MNIST sparsification experiment.
}
\label{tab:mnist_hparams}
\small
\setlength{\tabcolsep}{7pt}
\renewcommand{\arraystretch}{1.08}
\begin{tabular}{@{}lcc@{}}
\toprule
\textbf{Parameter}
& \textbf{Atomizer-IO}
& \textbf{ViT} \\
\midrule

Parameters (total)
& 7.40M
& 7.44M \\

Hidden / latent dimension
& 256
& 224 \\

Encoder / processor depth
& 1 cross-attn + 4 self-attn
& 12 transformer blocks \\

Attention heads
& 4 (cross) / 8 (self)
& 4 \\

MLP dimension
& 768
& 896 ($4\times$) \\

Positional encoding
& Relative (RoPE, self-attn)
& Absolute (learned) \\

Tokens / latents
& 1 spatial latent per retained pixel + 4 global
& 1 token per retained pixel + CLS \\

Attention dropout
& 0.05
& 0.05 \\

FF dropout
& 0.10
& 0.10 \\

Optimizer
& AdamW
& AdamW \\

Learning rate
& $1\times10^{-4}$
& $1\times10^{-4}$ \\

Weight decay
& $1\times10^{-3}$
& $1\times10^{-3}$ \\

Batch size
& 64
& 64 \\

Epochs
& 50
& 50 \\
\bottomrule
\end{tabular}
\end{table}

\subsection{LiDAR Training and Inference Protocol}
\label{app:lidar_protocol}

\paragraph{FRACTAL.}
We follow the RandLA-Net training protocol used in the FRACTAL benchmark for
the LiDAR stream. For each sample, all atoms generated from the co-registered
VHR image are retained, and $40{,}000$ LiDAR points are sampled and converted
to LiDAR atoms. The VHR and LiDAR atoms are concatenated into a single input
set before being passed to Atomizer-IO. No modality-specific processing branch
is used: both modalities are processed jointly through the same atomic encoder.

\paragraph{DALES.}
DALES scenes are cropped into $50\,\mathrm{m}\times50\,\mathrm{m}$ tiles for
training, and atom-to-latent assignments are precomputed for each tile. At
evaluation, each scene is covered by $50\,\mathrm{m}\times50\,\mathrm{m}$
sliding windows with a $25\,\mathrm{m}$ stride. For each window, the encoder
context contains up to the prescribed LiDAR-point budget, with subsampling when
necessary, while every point in the window is used as an output query.

Because adjacent windows overlap, a physical point may receive predictions from
multiple windows. We average the softmax class probabilities over all windows
covering that point and apply a single argmax afterwards, yielding exactly one
prediction per physical point. Predictions and ground-truth labels from all
scenes are then accumulated into a global confusion matrix from which per-class
IoU and mean IoU are computed.

Table~\ref{tab:dales_per_class} reports the resulting per-class IoU together
with the point-cloud baselines reported for DALES. Atomizer-IO reaches
$66.7\%$ mIoU, substantially above Perceiver IO in the same evaluation setting,
while remaining below the strongest task-specific point-cloud methods.

\begin{table*}[h]
\centering
\caption{
Per-class IoU on DALES. Values are reported as fractions. Baseline values are
those reported for the corresponding methods on DALES.
}
\label{tab:dales_per_class}
\small
\setlength{\tabcolsep}{4.5pt}
\renewcommand{\arraystretch}{1.05}
\begin{tabular}{@{}lccccccccc@{}}
\toprule
\textbf{Method}
& \textbf{Mean}
& \textbf{Ground}
& \textbf{Buildings}
& \textbf{Cars}
& \textbf{Trucks}
& \textbf{Poles}
& \textbf{Power lines}
& \textbf{Fences}
& \textbf{Vegetation} \\
\midrule
KPConv
& \textbf{0.811}
& 0.971
& \textbf{0.966}
& \textbf{0.853}
& \textbf{0.419}
& \textbf{0.750}
& \textbf{0.955}
& \textbf{0.635}
& \textbf{0.941} \\

PointNet++
& 0.683
& 0.941
& 0.891
& 0.754
& 0.303
& 0.400
& 0.799
& 0.462
& 0.912 \\

ConvPoint
& 0.674
& 0.969
& 0.963
& 0.755
& 0.217
& 0.403
& 0.867
& 0.296
& 0.919 \\

Atomizer-IO
& 0.667
& 0.962
& 0.941
& 0.670
& 0.095
& 0.549
& 0.829
& 0.395
& 0.895 \\

SuperPoint
& 0.606
& 0.947
& 0.934
& 0.629
& 0.187
& 0.285
& 0.652
& 0.336
& 0.879 \\

PointCNN
& 0.584
& \textbf{0.975}
& 0.957
& 0.406
& 0.048
& 0.576
& 0.267
& 0.526
& 0.917 \\

ShellNet
& 0.574
& 0.960
& 0.954
& 0.322
& 0.396
& 0.200
& 0.274
& 0.600
& 0.884 \\

Perceiver IO
& 0.487
& 0.944
& 0.895
& 0.362
& 0.032
& 0.144
& 0.453
& 0.228
& 0.840 \\
\bottomrule
\end{tabular}
\end{table*}
\newpage

\begin{table}[ht]
\centering
\caption{Notation used throughout the methodology.}
\label{app:notation}
\small
\setlength{\tabcolsep}{4.5pt}
\renewcommand{\arraystretch}{1.02}
\begin{tabular}{@{}lll@{}}
\toprule
\textbf{Stage} & \textbf{Symbol} & \textbf{Meaning} \\
\midrule

\multirow{9}{*}{Input}
& $N$ & Number of input atoms \\
& $v_i$ & Measured value of atom $i$ \\
& $J$ & Number of metadata fields of an atom \\
& $u_{i1},\dots,u_{iJ}$ & Metadata fields of atom $i$ \\
& $\phi_0(\cdot)$ & Encoder of the measured value \\
& $\phi_1(\cdot),\dots,\phi_J(\cdot)$ & Encoders of the metadata fields \\
& $\mathbf{f}_i$ & Token (feature vector) of atom $i$ \\
& $\mathbf{p}_i$ & Horizontal position of atom $i$, $\mathbf{p}_i\in\mathbb{R}^2$ \\
& $\mathcal{F}$ & Input set $\{(\mathbf{f}_i,\mathbf{p}_i)\}_{i=1}^{N}$ \\
\midrule

\multirow{9}{*}{Sampler}
& $\mathcal{C}$ & Context set $\{(c_i,\mathbf{p}_i)\}_{i=1}^{|\mathcal{C}|}$ \\
& $c_i$ & Feature of context element $i$ \\
& $\mathcal{Q}$ & Query set $\{(q_j,\mathbf{p}_j)\}_{j=1}^{|\mathcal{Q}|}$ \\
& $q_j$ & Feature of query element $j$ \\
& $\mathcal{A}_{\mathrm{enc}}$ & Encoder assignment: each context element to its nearest query \\
& $\mathcal{A}_{\mathrm{dec}}$ & Decoder assignment: $k$ nearest context elements per query \\
& $\mathcal{V}_j$ & Index set of the local receptive field of query $j$ \\
& $\boldsymbol{\delta}_{ij}$ & Relative offset $\mathbf{p}_i-\mathbf{p}_j$ of context element $i$ from query $j$ \\
& $\phi_\delta(\cdot)$ & Fourier encoding of a relative spatial offset \\
\midrule

\multirow{9}{*}{Encoder}
& $L$ & Number of spatial latents \\
& $D$ & Latent embedding dimension \\
& $\mathbf{p}_\ell$ & Spatial anchor of latent $\ell$ \\
& $\mathbf{h}_{\mathrm{init}}$ & Shared spatial-latent initialization \\
& $\mathcal{H}^{\mathrm{init}}$ & Initialized spatial latent set \\
& $\boldsymbol{\delta}_{i\ell}$ & Offset of atom $i$ relative to latent $\ell$ \\
& $\mathbf{z}_{i\ell}$ & Atom feature processed relative to latent $\ell$ \\
& $\mathbf{h}^{(0)}_\ell$ & Latent $\ell$ after encoder cross-attention \\
& $\mathcal{H}^{(0)}$ & Encoder output / processor input \\
\midrule

\multirow{10}{*}{Processor}
& $G$ & Number of global latents \\
& $B$ & Processor depth (number of blocks) \\
& $b$ & Processor block index, $b=0,\dots,B$ ($0$: before the first block) \\
& $\mathbf{g}^{(b)}_r$ & Global latent $r$ after processor block $b$ \\
& $\mathcal{G}^{(b)}$ & Global latent set after processor block $b$ \\
& $\mathbf{h}^{(b)}_\ell$ & Spatial latent $\ell$ after processor block $b$ \\
& $\mathcal{H}^{(b)}$ & Spatial latent set after processor block $b$ \\
& $\mathcal{H}^{(B)}$ & Final processed spatial latent set (decoder context) \\
& $s_p$ & Learned spatial RoPE scale \\
& $s_0$ & Reference scale used to initialize $s_p$ \\
\midrule

\multirow{8}{*}{Temporal}
& $T$ & Number of acquisition timesteps \\
& $t$ & Timestep index \\
& $\mathbf{h}^{(0)}_{t\ell}$ & Encoded latent $\ell$ at timestep $t$ \\
& $\mathcal{H}^{(0)}_t$ & Encoder output at timestep $t$ \\
& $\mathbf{h}^{(B)}_{t\ell}$ & Processed latent $\ell$ at timestep $t$ \\
& $\mathcal{H}^{(B)}_t$ & Processed spatial latent set at timestep $t$ \\
& $\bar{\mathbf{h}}_\ell$ & Temporally aggregated latent at anchor $\ell$ \\
& $\bar{\mathcal{H}}^{(B)}$ & Temporally aggregated set; replaces $\mathcal{H}^{(B)}$ as decoder context \\
\midrule

\multirow{9}{*}{Decoder}
& $M$ & Number of output queries \\
& $\mathbf{p}_j$ & Position of output query $j$ \\
& $k$ & Number of nearest spatial latents gathered per output query \\
& $\mathbf{q}_{\mathrm{init}}$ & Shared output-query initialization \\
& $\mathcal{Q}^{\mathrm{init}}$ & Initialized output-query set \\
& $\boldsymbol{\delta}_{\ell j}$ & Offset of latent $\ell$ relative to output query $j$ \\
& $\mathcal{S}_j$ & Indices of input atoms co-located with output query $j$ \\
& $\tilde{\mathbf{q}}_j$ & Query conditioned on co-located observations \\
& $\mathbf{q}_j$ & Decoded representation at output query $j$ \\
\midrule

\multirow{2}{*}{Budgets}
& $\kappa$ & Target number of atoms per spatial latent \\
& $m$ & Maximum number of atoms sampled per Voronoi cell \\
\bottomrule

\end{tabular}
\end{table}
\end{document}